\PassOptionsToPackage{table}{xcolor}
\documentclass[]{alaya}
\usepackage{makecell}
\usepackage{wrapfig}
\usepackage{tabularx}
\usepackage{textcomp}
\usepackage{stfloats}
\usepackage{url}
\usepackage{verbatim}
\usepackage{titlesec}
\usepackage{adjustbox}
\usepackage{multirow}
\usepackage{pifont}
\usepackage[sc]{mathpazo}
\usepackage{tikz}
\usepackage{comment}
\usepackage{amsmath,amssymb}
\usepackage{colortbl}
\usepackage{natbib}
\usepackage{color}
\usepackage{booktabs} 
\usepackage{hyperref}
\usepackage{graphicx}
\usepackage{subcaption}
\RequirePackage{xspace}
\makeatletter
\DeclareRobustCommand\onedot{\futurelet\@let@token\@onedot}
\def\@onedot{\ifx\@let@token.\else.\null\fi\xspace}
\usepackage[most]{tcolorbox}
\usepackage{xcolor}
\usepackage{array}
\usepackage{tabularx}
\usepackage{siunitx} 
\usepackage{makecell}
\usepackage[table]{xcolor}
\usepackage{caption}
\definecolor{headerpurple}{HTML}{d8d2fc}
\definecolor{rowgray}{gray}{0.95}

\makeatother

\definecolor{adptorange}{RGB}{248, 205, 172}
\definecolor{cmpblue}{RGB}{189, 215, 238}
\definecolor{cmpblue}{RGB}{189, 215, 238}

\definecolor{our_red}{RGB}{232,157,160}
\definecolor{our_blue}{RGB}{136,206,230}
\definecolor{our_orange}{RGB}{246,200,168}
\definecolor{our_green}{RGB}{178,211,164}

\definecolor{attn_code0}{RGB}{247,215,200}
\definecolor{attn_code1}{RGB}{238,169,139}
\definecolor{mlp_code0}{RGB}{204,201,221}
\definecolor{mlp_code1}{RGB}{102,95,153}
\definecolor{mygray}{HTML}{f0f0f0}

\definecolor{token_blue}{RGB}{84, 120, 140}

\definecolor{aicolor}{RGB}{0,0,0}
\DeclareRobustCommand{\tocheck}[1]{{\color{aicolor}#1}}
\definecolor{takeawaybg}{RGB}{240,244,250}
\definecolor{takeawayframe}{RGB}{31,95,191}
\newtcolorbox{takeawaybox}{enhanced,breakable=false,colback=takeawaybg,colframe=takeawaybg,boxrule=0pt,leftrule=2.5pt,colbacktitle=takeawaybg,borderline west={2.5pt}{0pt}{takeawayframe},arc=1pt,outer arc=1pt,left=6pt,right=6pt,top=3pt,bottom=3pt,before skip=6pt,after skip=4pt}
\newcommand{\takeaway}[2]{\begin{takeawaybox}\textbf{\textcolor{takeawayframe}{#1}}\quad #2\end{takeawaybox}}

\usepackage{pifont}
\usepackage{bbding}
\usepackage{fontawesome}

\newcommand{\cmark}{\ding{51}}
\newcommand{\best}[1]{{\bfseries #1}}
\newcommand{\second}[1]{\underline{#1}}
\newcommand{\ua}{\ensuremath{\uparrow}}
\newcommand{\da}{\ensuremath{\downarrow}}
\makeatletter
\newcommand{\inpicon}[1]{\csname inp@icon@#1\endcsname}
\def\inp@icon@FF{\faPictureO}       % first frame
\def\inp@icon@UFF{{\sbox\z@{\faPictureO}\makebox[\wd\z@][c]{\raisebox{0pt}[\ht\z@][\dp\z@]{\rotatebox[origin=c]{-20}{\faPictureO}}}}} % unverified first frame: tilted, same footprint as FF
\def\inp@icon@C{\faVideoCamera}     % camera pose
\def\inp@icon@D{\faDatabase}        % depth
\def\inp@icon@CS{\faPaintBrush}     % class-level segmentation
\def\inp@icon@M{\faAdjust}           % mixed render
\def\inp@icon@P{\faCube}             % 3D proxy
\def\inp@icon@B{\faSquareO}         % 3D bounding boxes
\def\inp@icon@AB{\faObjectGroup}    % all-entity boxes
\newcommand{\inpvocab}{FF,C,D,CS,M,P,B}
\newcommand{\inp}[1]{{\def\inp@on{}\forcsvlist{\listadd\inp@on}{#1}\def\inp@sep{}%
  \expandafter\forcsvlist\expandafter\inp@one\expandafter{\inpvocab}}}
\newcommand{\inp@one}[1]{\inp@sep\def\inp@sep{\hspace{0.9pt}}%
  \ifstrequal{#1}{FF}{\xifinlist{UFF}{\inp@on}{\inpicon{UFF}}{\inp@draw{#1}}}{\inp@draw{#1}}}
\newcommand{\inp@draw}[1]{\xifinlist{#1}{\inp@on}{\inpicon{#1}}{\textcolor{black!18}{\inpicon{#1}}}}
\newcommand{\inpkey}[2]{\inpicon{#1}\,=\,#2}
\makeatother
\usepackage{float}

\newlength\savewidth
\newcommand{\tablestyle}[2]{\setlength{\tabcolsep}{#1}\renewcommand{\arraystretch}{#2}\centering\footnotesize}

\newcolumntype{x}[1]{>{\centering\arraybackslash}p{#1pt}}
\newcolumntype{y}[1]{>{\raggedright\arraybackslash}p{#1pt}}
\newcolumntype{z}[1]{>{\raggedleft\arraybackslash}p{#1pt}}

\renewcommand{\paragraph}[1]{\vspace{1mm}\noindent\textbf{#1}}
\usepackage{colortbl}
\usepackage{xcolor}
\usepackage{wrapfig}

\usepackage{booktabs}
\usepackage{multirow}
\usepackage{graphicx}
\usepackage{amssymb}
\usepackage{makecell}
\usepackage{array}

\renewcommand{\arraystretch}{1.15}

\renewcommand{\paragraph}[1]{\vspace{1.25mm}\noindent\textbf{#1}}

\usepackage{algorithm}
\usepackage{listings}

\definecolor{codeblue}{rgb}{0.25, 0.5, 0.5}
\definecolor{codekw}{rgb}{0.35, 0.35, 0.75}
\lstdefinestyle{Pytorch}{
    language = Python,
    backgroundcolor = \color{white},
    basicstyle = \fontsize{9pt}{8pt}\selectfont\ttfamily\bfseries,
    columns = fullflexible,
    aboveskip=1pt,
    belowskip=1pt,
    breaklines = true,
    captionpos = b,
    commentstyle = \color{codeblue},
    keywordstyle = \color{codekw},
}

\definecolor{green}{HTML}{009000}
\definecolor{red}{HTML}{ea4335}

\title{\tocheck{\raisebox{-0.03em}{\includegraphics[height=0.77em]{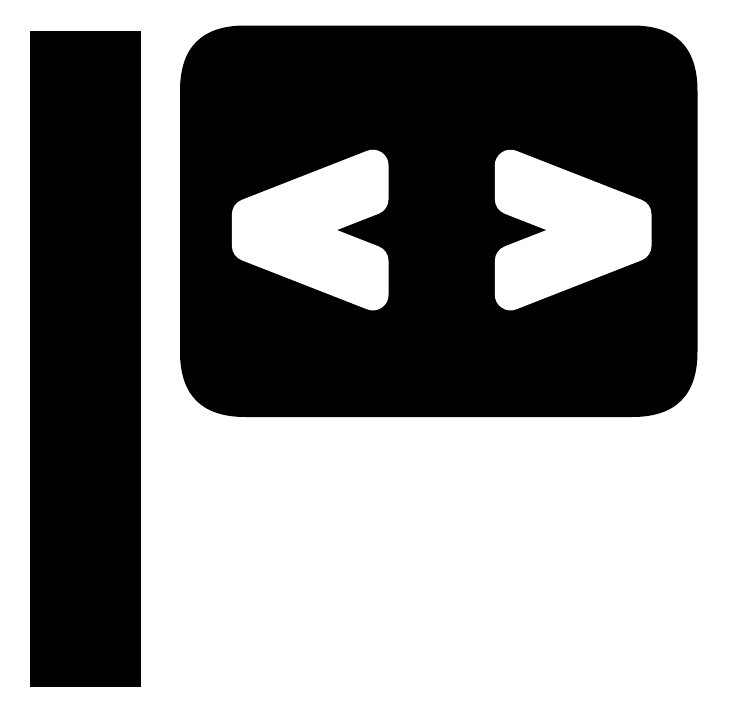}}\kern0.04emROWBench: Do Video Models Render\\What the Program Specifies?}}
\author[\triangle,\star]{Zheng-Hui Huang}
\author[\triangle,\star]{Guixu Lin}
\author[\triangle]{Yu-Ju Tsai}
\author[]{Jian-Kai Zhu}
\author[\triangle]{Fengbo Lan}
\author[]{Yu-Lun Liu}
\author[]{Yung-Yu Chuang}
\author[\triangle,\dagger]{Kaipeng Zhang}
\author[\triangle,\dagger]{Zhixiang Wang}

\affiliation[\triangle]{Alaya Lab}
\abstract{% Programmable world models separate executable world dynamics from visual generation, yet correct state evolution does not guarantee faithful visual realization. 
Programmable world models separate executable dynamics from visual generation, offering a promising foundation for next-generation game engines. However, their visual adherence to explicit rules and interactions remains insufficiently evaluated. \tocheck{Existing benchmarks assess visual quality, controllability, and instruction or physical adherence, but rarely test fidelity to fine-grained, program-specified world events.} We introduce \textbf{PROWBench}, comprising 170 programmatically constructed episodes and 600 proxy videos covering diverse scenes and interactions. \tocheck{PROWBench logs entity states and timestamped events, including those outside the camera's field of view, as replayable world records, from which it renders synchronized views and proxy representations. This enables generated videos to be checked against the observable consequences of program execution. An extensible framework constructs scenes, controls behaviors, and can render each camera view in different representations, such as coarse 3D, and bounding boxes.} The benchmark covers first- and third-person perspectives, with synchronized multi-view observations available for a subset of episodes. \tocheck{Grounded in these records, PROWBench evaluates entity control, long-horizon memory, and, with two VLM-based metrics, \textbf{Logic-Render Alignment} and \textbf{Interaction Success Rate}, adherence to the prescribed timeline and the visual realization of timestamped engine-recorded events.}

}

\project{\url{https://alaya-lab.github.io/PROWBench}}
\code{\url{https://github.com/AlayaLab/PROWBench}}
\correspondence{Zhixiang Wang (Project Lead), Kaipeng Zhang}
\contributionmark{$\star$ denotes equal contributions}
\date{\today}

\begin{document}
\maketitle
% \pagestyle{fancy}

% %在这里添加你的 Teaser 图片
% \teaser{
\begin{figure}[H]
  \centering
  \includegraphics[width=\linewidth]{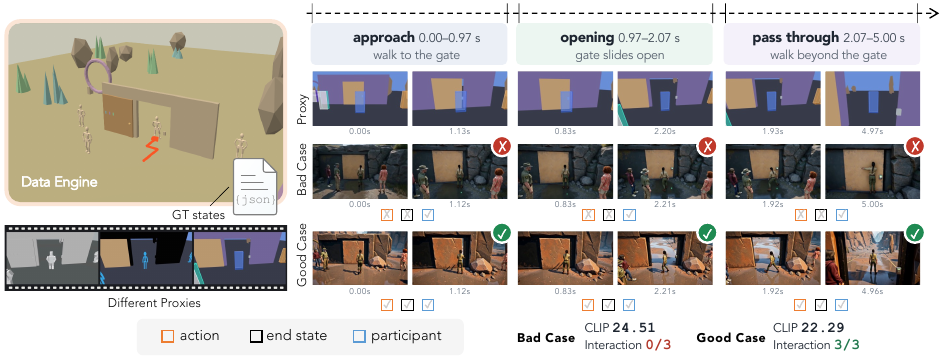}
  \caption{\textbf{Overview of our extensible framework and benchmark.}
  % Our extensible framework pairs visual proxies with ground-truth world records to evaluate how faithfully generated videos realize prescribed states, rules, and interactions. 
  % Without the GT states, evaluation would be problematic.
  Our extensible framework pairs visual proxies with ground-truth world records, enabling reliable evaluation of how faithfully generated videos realize prescribed states, rules, and interactions. This assessment would otherwise be difficult. Here, we compare the CLIP score with our proposed interaction metric. CLIP assigns a higher score to the clearly inferior case, whereas one of our metric correctly reflects whether the prescribed interaction is faithfully executed.
  % \zhixiangc{This figure or the figure after it needs to show why the previous benchmarks are not suitable to eval PWM...}
  }
  \label{fig:teaser}
  % \vspace{-2mm}
\end{figure}
% }
% % % teasor

\section{Introduction}
\label{sec:introduction}

% \TODO{Unify the terminology across the paper. There is no terminology for white proxy...}

Recent advances in video world models have enabled increasingly realistic, open-ended interactive environments. However, maintaining persistent world states and enforcing user-defined rules remain challenging when entities, attributes, and interaction outcomes are represented implicitly through visual histories or latent context. Programmable World Model and Code World Model address these limitations by separating world-state evolution from visual generation~\citep{huang2026programmableworldmodel, chen2026code}: coding agents and executable programs maintain states and interaction rules, while video models generate observations conditioned on structured representations of the world. This separation combines explicit state control with visual generation, but correct program execution does not guarantee faithful visual realization. A generated video may appear plausible while violating a rule or failing to depict a specified interaction. This raises a central evaluation question: \emph{can video models faithfully render the scene structure, rules, and interactions of a programmable world?}

Existing benchmarks evaluate video quality~\citep{huang2024vbench}, controllability and dynamics~\citep{duan2025worldscore}, instruction following and physical adherence~\citep{li2025worldmodelbench}, and interactive control and consistency~\citep{ye2026mind,shang2026worldarena,wu2026omniworldbench,xu2026worldmark,ying2026wbench}. However, as summarized in Table~\ref{tab:dataset_comparison}, they rarely test fidelity to fine-grained, program-specified world events: without replayable records of entity states and timestamped events, a generated video can be judged for plausibility but not against what the program executed. None of them renders the same recorded scene in paired proxy representations, and synchronized views are rare, although both are needed to isolate the effects of representation and viewpoint while holding scene content and actions fixed.

To address these gaps, we introduce \textbf{PROWBench}, a benchmark for evaluating the visual realization of programmable worlds. Figure~\ref{fig:teaser} provides an overview of our extensible framework and benchmark. PROWBench combines a programmatically constructed dataset with complementary metrics for logic adherence and interaction realization. The data engine is available for extension. It logs entity states and timestamped events, including those outside the camera's field of view, as replayable world records and renders them into synchronized views and proxy representations, so that generated videos can be checked against the observable consequences of program execution.

PROWBench contains 170 dynamic scene episodes covering diverse environments and interactions. Each episode currently provides three synchronized representations (coarse 3D, semantic proxy, and colored oriented bounding boxes) from shared state, action, and interaction records, enabling controlled comparisons across representations. Some episodes additionally provide synchronized multi-view observations for cross-view comparisons. Grounded in these records, PROWBench evaluates entity control and the long-horizon memory of entities that leave and re-enter the view, and introduces two VLM-based metrics: \textbf{Logic-Render Alignment} assesses adherence to the prescribed timeline, while \textbf{Interaction Success Rate} checks whether recorded events are visually realized at the corresponding times using engine logs as ground truth. An extensible programmatic framework separates scene construction, behavior control, state recording, and rendering, allowing researchers to expand scene layouts, actions, and representations while generating aligned observations from shared state records.

We evaluate state-of-the-art models with the inputs produced by this data engine: coarse 3D, semantic-proxy, and colored-OBB renders of the recorded states, together with text prompts obtained by converting the engine's recorded events. Our experiments assess their logic adherence and interaction realization across viewpoints and representation settings. \tocheck{Surprisingly, with appropriate prompts, existing general-purpose video generation models such as MiniMax-H3~\citep{minimaxh3}, without task-specific training, follow the prescribed entity trajectories and event logic well, and perform on par with camera-conditioned world models in camera following without explicit camera injection. Additional task- or domain-specific training can still improve several aspects, such as entity placement and the realization of domain-specific events. The proxy itself also matters: outputs sometimes follow the input geometry too closely and inherit the coarse shapes of low-poly proxies, while box proxies leave open which described appearance belongs to which entity, so designing a good proxy remains an open problem.} Long-video generation remains challenging in terms of both camera following and \tocheck{entity persistence}, while direct multi-view generation still yields unsatisfactory results.

Our contributions are threefold:
\begin{itemize}
    \item \textbf{Benchmark.}
    We introduce PROWBench, comprising 170 episodes and 600 synchronized proxy videos covering multiple perspectives, visual representations, and diverse interactions, with shared world-state and action records.
    \item \textbf{Metrics.}
    \tocheck{We evaluate entity control and long-horizon memory against the recorded world states, and propose Logic-Render Alignment and Interaction Success Rate to assess whether generated videos follow the prescribed timeline and visually realize engine-recorded interaction events.}
    \item \textbf{Extensible framework.}
    We provide a reusable programmatic generation framework that supports extensions to scene layouts, actions, and visual representations, enabling aligned multi-view and multi-representation data generation from shared state records.
\end{itemize}

\section{Related Work}
\label{sec:related_work}

\subsection{Video World Models and Programmable World States}

Video world models predict future observations from visual context and actions. Genie~\citep{bruce2024genie} learns controllable environments from unlabeled videos, while GameNGen~\citep{valevski2025gamengen} enables real-time action-conditioned game generation. GameFactory~\citep{yu2025gamefactory}, Matrix-Game~\citep{zhang2025matrixgame}, Yume1.5~\citep{mao2026yume15}, and LingBot-World~\citep{robbyant2026lingbotworld} advance generalization, interaction, and long-horizon consistency. Memory-enhanced WorldPlay~\citep{sun2025worldplay}, Matrix-Game 3.0~\citep{wang2026matrixgame3}, and AlayaWorld~\citep{team2026alayaworldv11} improve persistence through spatial-context retrieval or reconstruction.

While these models infer states from visual history and memory, programmable world models externalize state evolution. Code World Model~\citep{chen2026code} uses a coding agent to maintain executable states and interaction logic, while Programmable World Model~\citep{huang2026programmableworldmodel} feeds persistent structured states to a generative renderer. Other approaches similarly separate state or dynamics from appearance~\citep{meng2026marionette,lin2026stateplay,cai2026mass}. PROWBench evaluates whether generated observations faithfully realize these explicit states and interactions.

\subsection{Structured Representations for Visual Generation}

Structured conditions offer direct spatial control through depth, trajectories, bounding boxes, masks, and camera motion~\citep{esser2023structure,wang2024boximator,li2025trackdiffusion,wang2024motionctrl,luo2025ctrlv,jain2024peekaboo,li2025magicmotion}. Geometry-aware methods incorporate 3D boxes, bird's-eye-view layouts, world volumes, and structured traffic scenes~\citep{gao2024magicdrive,li2024drivingdiffusion,wen2024panacea,lu2024wovogen,wang2024drivedreamer,wang2024drivewm,wang2025cinemaster}.
Generative rendering uses explicit geometry or renderer-derived signals: Coarse-to-Real~\citep{gomez2026coarse} and VideoFrom3D~\citep{kim2025videofrom3d} use coarse 3D scenes, while Diffusion Renderer~\citep{liang2025diffusionrenderer}, AlayaRenderer~\citep{huang2026generativeworldrenderer}, its real-time extension~\citep{lin2026generativeworldrendererspeed}, and LiVER~\citep{cai2026liver} use denser rendering cues. These approaches span sparse object controls to dense geometric representations.
Comparisons across representations are often confounded by differences in models, datasets, and tasks. PROWBench derives paired proxy representations from the same recorded world-state trajectory and camera configuration, isolating representation effects on visual realization under fixed scene evolution.

\subsection{World Model Benchmarks and Evaluation}

VBench~\citep{huang2024vbench} and EvalCrafter~\citep{liu2024evalcrafter} assess video quality, temporal consistency, motion, and semantic alignment. WorldSimBench~\citep{qin2025worldsimbench}, WorldModelBench~\citep{li2025worldmodelbench}, and WorldScore~\citep{duan2025worldscore} extend evaluation to physical plausibility, controllability, and world dynamics. MIND~\citep{ye2026mind}, WorldMark~\citep{xu2026worldmark}, Omni-WorldBench~\citep{wu2026omniworldbench}, WorldArena~\citep{shang2026worldarena}, and iWorld-Bench~\citep{iworldbench2026} further target interactive behavior, memory, and long-horizon consistency.

WBench~\citep{ying2026wbench} evaluates multi-turn navigation, subject actions, event editing, and perspective switching for instruction adherence, consistency, and physical plausibility. These benchmarks generally assess complete pipelines, coupling errors in action understanding, state prediction, memory, dynamics, and rendering.

PROWBench isolates rendering fidelity using engine-recorded frame-level states, rules, and timestamped interactions, enabling comparisons across synchronized perspectives and proxy representations under the same trajectory. Logic-Renderer Alignment measures whether specified rules are visibly realized, while Interaction measures whether recorded events appear at the corresponding times.
\section{Benchmark}
\begin{figure}
    \centering
    \includegraphics[width=\linewidth]{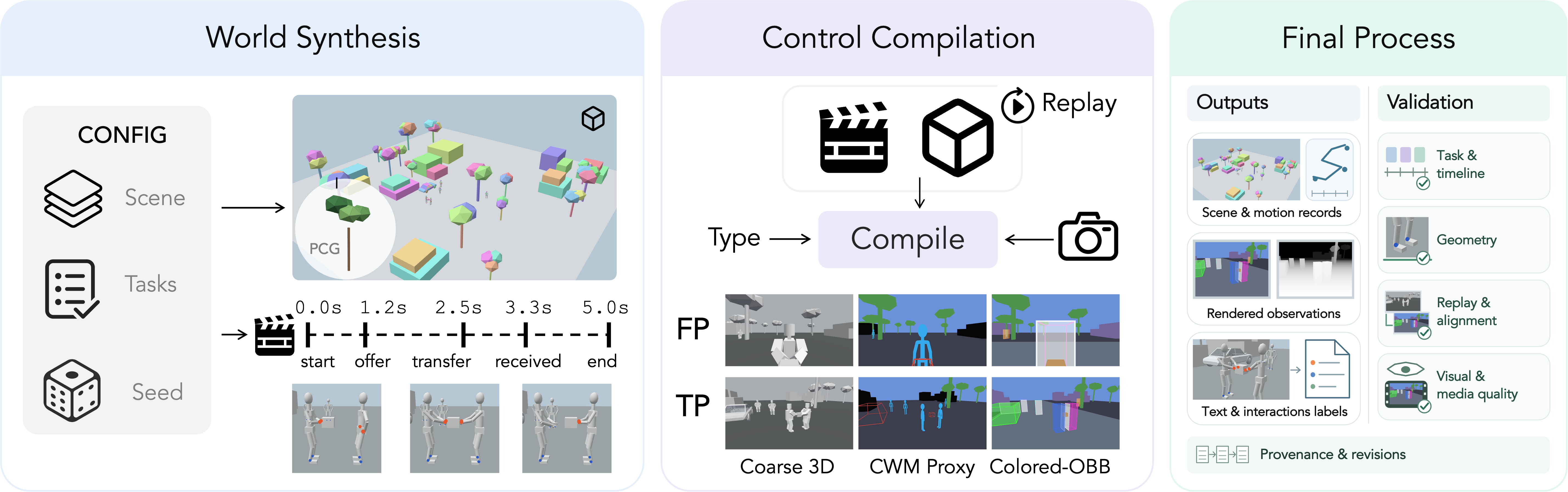}
    \caption{\textbf{Overview of the world generation pipeline.}
    World synthesis generates replayable trajectories, control compilation produces synchronized conditioning inputs across viewpoints and representations, and the final process validates and exports videos with structured annotations.}
    % \vspace{-4mm}
    \label{fig:pipeline}
\end{figure}

\begin{figure}[tb]
    \centering
    \includegraphics[width=\linewidth]{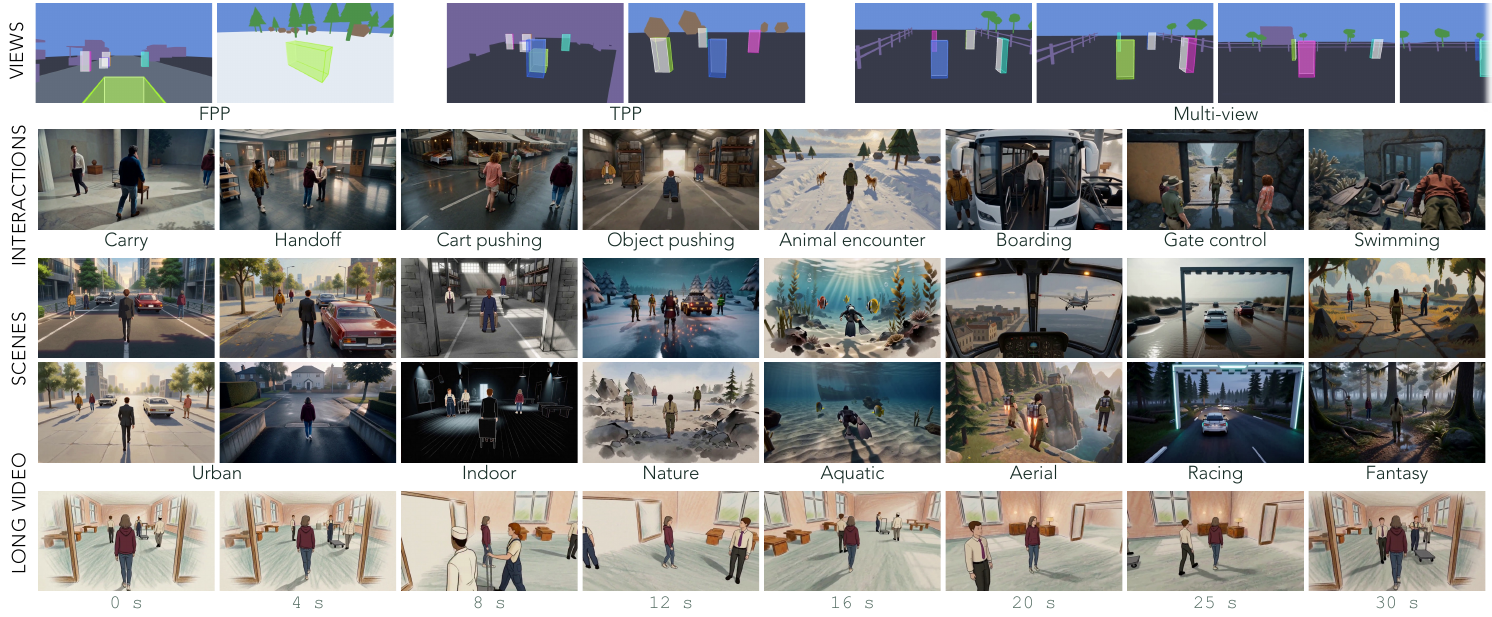}
        \caption{\textbf{Diversity of PROWBench.} PROWBench provides diverse camera perspectives, interactions, scene domains, 
        and long-horizon videos.
        % and 30-second sequences for long-horizon evaluation.
        }
        \label{fig:prowbench_data}
        % \vspace{-4mm}
    \end{figure}

We construct the PROWBench dataset using a programmable world generation pipeline that produces aligned conditioning inputs and evaluation records. Figure~\ref{fig:pipeline} outlines the pipeline, while Figure~\ref{fig:prowbench_data} illustrates the dataset's content and diversity. We describe the pipeline and evaluation suite below, with detailed dataset statistics provided in Section~\ref{sec:data_statistics}.

% \TODO{Figure~\ref{fig:pipeline}: show clearly the artifacts that running our released code produces (e.g., scene/world records, camera and entity trajectories, event logs, synchronized proxy videos in each representation, depth/segmentation, prompts and engine annotations), so readers see what the framework outputs, not only the pipeline stages.}

\subsection{World Generation Pipeline}
\label{sec:generation_system}

We introduce a pipeline for constructing programmable worlds and producing structured conditioning inputs and evaluation annotations for generative video models. The pipeline provides three complementary capabilities: \emph{world synthesis}, \emph{control compilation}, and \emph{final process}. Together, these capabilities produce replayable world trajectories, synchronized conditioning inputs, and structured evaluation records.

% \noindent \textbf{World Synthesis.}
\noindent \textbf{World Synthesis.}
The pipeline constructs dynamic worlds from configurations specifying the environment, entity layout, participating actors, task controllers, random seeds, and execution duration. Procedural routines generate terrain, buildings, roads, interiors, and interaction objects, together with collision surfaces, collision constraints, and interaction anchors. Entities have persistent episode-level identifiers, semantic categories, component hierarchies, and task-relevant attributes. Random seeds control variations in scene layouts, motion paths, speeds, and animation phases.

Specialized controllers coordinate articulated poses, object trajectories, and participant interactions through procedural animation and motion models. They support three broad categories of behavior. Interaction controllers organize tasks into explicit phases: for example, carrying proceeds through approaching, grasping, lifting, transporting, and releasing an object, with corresponding updates to hand targets and object ownership. Activity and locomotion controllers support animal behaviors and simplified flight and swimming dynamics. Driving controllers provide closed-loop control of vehicle motion. During execution, these controllers report action phases, participating entities, and relationship changes for recording and validation. External agents can query entities, submit supported actions or plans, advance the world by fixed ticks, and inspect execution feedback.

Each execution produces a structured episode record containing shared scene geometry, entity definitions and hierarchies, initial conditions, and temporal records of entity transforms, articulated poses, visibility, and appearance changes. Controller events and available input traces accompany these records, with explicit temporal mappings linking world execution, saved states, and video frames. Each saved state can be reconstructed independently from the shared scene description and its corresponding dynamic record. Configurations, random seeds, and source information are retained for provenance tracking.

\noindent \textbf{Control Compilation.}
The pipeline compiles recorded world states and camera trajectories into synchronized visual conditioning inputs for generative video models.

Third-person cameras follow task-dependent trajectories, while first-person cameras are derived from recorded head positions, character or vehicle orientations, and task-specific gaze information. The observing character's display geometry is hidden in its first-person view, while its entity and articulated state remain in the shared world record. Final camera trajectories include intrinsics, camera-to-world and world-to-camera matrices, projection parameters, and clipping planes. Additional cameras can be associated with different participants to provide multiple perspectives of the same episode.

Rendering adapters map shared world records to conditioning representations. Our current implementation provides three representations. \emph{Coarse 3D representation (Coarse 3D)} uses neutral materials to preserve scene shape, object silhouettes, and articulated motion. \emph{Semantic proxy (CWM proxy)} combines semantically colored environments with simplified entity proxies, retaining global trajectories, orientations, and task-relevant posture while reducing local articulation and shape detail. \emph{Colored oriented bounding boxes (Colored-OBB)} represent selected entities through spatial envelopes, stable instance colors, and directional cues, with a semi-transparent white face marking the front of each entity, while retaining geometry for environmental structures and simple objects. OBB bounds are updated over time, while instance colors remain consistent across frames and views.

Rendering adapters specify entity selection, proxy templates, transparency, occlusion, and clipping behavior. Given shared scene geometry $G$, recorded world state $S_i$, camera parameters $C_i^v$ for view $v$, and representation $r$, a conditioning frame is generated as $I_i^{v,r} = \mathcal{R}_r(G, S_i, C_i^v)$,
% \begin{equation}
%     I_i^{v,r} = \mathcal{R}_r(G, S_i, C_i^v),
%     \label{eq:paired_rendering}
% \end{equation}
where $\mathcal{R}_r$ is the representation-specific rendering adapter. Different views observe the same saved state, and representations within each view share camera parameters. Additional camera policies and rendering adapters can reuse recorded episodes without rerunning behavior controllers, provided the required information is available in the world records, whereas new behaviors produce new recorded executions. This separation preserves temporal alignment and enables controlled comparisons across representations and viewpoints while keeping the underlying world evolution fixed.

\noindent \textbf{Final Process.}
The pipeline exports three categories of outputs: scene and motion records, rendered observations, and text and interaction labels. Scene and motion records support replay and include scene geometry, entity identities and semantic categories, spatial and articulated states, joint poses, bounding volumes, and entity and camera trajectories with associated camera parameters. Rendered observations comprise synchronized proxy videos across viewpoints and representations, accompanied by metric depth interpreted with respect to its recorded geometry. Text and interaction labels describe action phases, participant relationships, and timestamped events, together with accompanying textual descriptions and contact targets and residuals. 
% The controller or inverse-kinematics provenance of contact targets and residuals is retained to support their interpretation alongside independent geometric checks.

Validation comprises four complementary components. Task and timeline checks assess scenario-specific phases, terminal conditions or state validity, and recording coverage. They also verify task-level interaction labels against the applicable annotation policy and required participant categories. Geometric checks evaluate penetration, support, and contact consistency at saved states using canonical geometry. Replay and alignment checks assess state reconstruction, camera alignment, and consistency with the selected representation. They further establish correspondence among videos, depth, and state labels using source records, frame indices, camera information, and timestamps. Visual and media quality checks combine first-frame visibility measurements based on canonical-geometry pixel counts and view-frustum tests with documented visual inspection of first frames and selected temporal samples. These inspections assess subject visibility and observation readability, while media decoding and hash checks assess output integrity. Visibility measurements and visual reviews are interpreted within the scope of the recorded frames and geometry.

To preserve provenance and revision history, failed attempts and their available diagnostics are retained separately from accepted episodes. For each batch-specific revision, affected outputs are backed up, changes to world records, camera trajectories, or representation policies are documented, and the corresponding media and metadata are updated together. Hash checks verify the integrity of unchanged source records where applicable. Sample metadata and source hashes link released observations to their source records, rendering configurations, and verification reports.

\subsection{Evaluation Suite}
\label{sec:eval}
We evaluate five complementary capabilities: controllability, visual and temporal quality, logic and state alignment, multi-view consistency, and long-horizon memory. Engine-recorded trajectories, events, and states provide references for evaluating adherence to the prescribed world evolution, while established perceptual metrics assess video quality.

\noindent \textbf{Controllability.}
We evaluate whether entities and the camera follow their prescribed trajectories. For entity control, we consider dynamic and interaction-relevant entities visible in the reference view, excluding terrain, effects, and scene geometry. Generated entities are localized with SAM~3~\citep{carion2026sam} from semantic-label prompts and matched to engine-projected references by label-compatible Hungarian matching. \textbf{BBox IoU} measures spatial alignment over the visible trajectory, with missed detections assigned zero IoU, while \textbf{Center Error} measures normalized center displacement conditional on detection. We report clip-level averages over evaluable entities; full visibility, matching, and aggregation details are provided in Appendix~\ref{appendix_metric}.

\noindent \textbf{Interaction Success Rate.}
Global similarity alone does not determine whether a prescribed interaction is successfully executed. We therefore evaluate registered engine events, each defined by its participants, temporal window, action, and expected end state. For each event, Qwen3.6-27B~\citep{qwen3.6-27b} is provided with eight frames sampled around the event window, including both full-frame context and an engine-localized crop, and independently judges the prescribed action, expected end state, and participant presence. An event is considered successful only when the response is valid and both the action and end state are judged present. 
For a generated scene-camera sample $s$ with event set $\mathcal{E}_s$, we define
\begin{equation}
    \mathrm{ISR}_s = \frac{1}{|\mathcal{E}_s|}\sum_{e\in\mathcal{E}_s}
    \mathbf{1}[\text{valid response, action=yes, end state=yes}].
    \label{eq:interaction_success}
\end{equation}
Table entries average $\mathrm{ISR}_s$ equally across samples, preventing samples with more annotated events from receiving greater weight. 

We additionally report $\mathrm{gISR}_s$, an IoU-weighted variant of ISR that jointly reflects interaction success and entity placement:
\begin{equation}
    \mathrm{gISR}_s = \mathrm{ISR}_s \cdot \mathrm{IoU}_s,
\end{equation}
where $\mathrm{IoU}_s$ is the clip-level BBox IoU of sample $s$. Table entries average $\mathrm{gISR}_s$ equally across samples.

\noindent \textbf{Logic-Render Alignment.} 
Logic-Render Alignment measures whether each segment of the action timeline in the engine-derived prompt is visibly depicted within its time window.
For each annotated timeline segment, the judge receives twelve frames sampled from the corresponding temporal window, together with the expected event description and neighboring descriptions as context. The metric evaluates whether the action prescribed by the engine logic is visibly realized within the corresponding segment. For a sample $s$ with annotated timeline segments $\mathcal{T}_s$, we define
\begin{equation}
    \mathrm{LRA}_s = \frac{1}{|\mathcal{T}_s|}\sum_{\tau\in\mathcal{T}_s}
    \mathbf{1}[\text{prescribed action in segment }\tau\text{ is judged present}],
    \label{eq:logic_render_alignment}
\end{equation}
so that $\mathrm{LRA}_s$ is the fraction of timeline segments whose prescribed actions are visually consistent with the engine-defined logic.

Similarly, we report $\mathrm{gLRA}_s$, an IoU-weighted variant of LRA that jointly reflects timeline adherence and entity placement:
\begin{equation}
    \mathrm{gLRA}_s = \mathrm{LRA}_s \cdot \mathrm{IoU}_s.
\end{equation}
Table entries average $\mathrm{gLRA}_s$ equally across samples.

Figure~\ref{fig:metric_illustration} illustrates the distinction between the two behavioral metrics. ISR evaluates whether a registered interaction is successfully completed, requiring both the prescribed action and expected end state, whereas LRA evaluates whether the generated video realizes the engine-defined event logic by depicting each prescribed action in its corresponding temporal segment.

\noindent \textbf{Memory over Long Horizons.}
For the 30-second evaluation, we test whether entities that leave the field of view reappear at the location and in the state prescribed by the engine. We define an \emph{absence event} when an entity becomes invisible and later re-enters the view. \textbf{Reappearance IoU} measures spatial recovery after re-entry by comparing the reference box with label-compatible detections, with missed detections assigned zero IoU. \textbf{State Persistence} measures whether a returning person preserves the prescribed locomotion state, classified as \emph{walking} or \emph{standing} and judged from post-return frames by Qwen3.6-27B~\citep{qwen3.6-27b}. Both metrics are averaged first within scenes and then across scenes with valid events. 
%This judge has not been calibrated on engine renders of this set.
%\TODO{}

Additional metric definitions, implementation details, and evaluation protocols are provided in Appendix~\ref{appendix_metric}.

\subsection{\texorpdfstring{\tocheck{Benchmark Tracks}}{Benchmark Tracks}}
\label{sec:tracks}

\tocheck{PROWBench organizes its evaluation cases into two tracks with four settings. Each setting uses its own evaluation set, and the scenes of the two main-track settings are disjoint.}

\noindent \tocheck{\textbf{Main Track.}} \tocheck{The main track evaluates controlled generation of five-second clips in two settings that differ in the appearance reference and use disjoint scenes.} When the agent already has an appearance reference, such as a concept image or a prescribed character design, it can supply it as the RGB first frame. \tocheck{The Verified-FF setting evaluates this case with 40 scene-camera pairs (31 third-person follow, 9 first-person; top of Table~\ref{tab:dv200_firstframe}). Its RGB first frame is part of the benchmark: it is generated with MiniMax-H3 from the colored-OBB proxy and the scene description and checked against the recorded person count, positions, facing and layout (Appendix~\ref{app:firstframe}), and every method that accepts a first frame receives this identical image.} In procedural game-world construction, the agent usually has no ready-made RGB appearance reference. \tocheck{The Unverified-FF setting therefore prescribes only the world state, the text, and the control signals, and asks whether models can jointly solve entity control, appearance generation, and identity binding. It contains 90 pairs (78 third-person follow, 12 first-person; bottom of Table~\ref{tab:dv200_nofirstframe}), and appearance must be determined by each model from the text and the control signal.} The curation of this set used MiniMax-H3 outputs, so results involving MiniMax-H3 should be read with this potential bias in mind.

\noindent \tocheck{\textbf{Challenge Track.}} \tocheck{The challenge track tests whether short-clip control extends to two harder settings.} Control in short clips does not guarantee that a model maintains one world state over longer horizons, in which entities leave the view, re-enter later and keep their state and motion across generation windows. \tocheck{The long-horizon setting (Table~\ref{tab:long_horizon}) contains ten 30-second recordings, one camera each.} In a multi-player game, each player observes the same world through their own camera. Even when each view is generated independently, the views must agree on where each participant is, who they are, what they look like, and the scene structure. \tocheck{The multi-view setting (Table~\ref{tab:multiview_crossview}) contains} ten four-player scenes, each with four synchronized third-person follow views, one per player, generated independently.

\section{Experiments}
\label{sec:experiments}

Our experiments address a central question: can current video generation and video world models faithfully render a prescribed dynamic world, rather than merely produce visually plausible videos? For game worlds, visual quality alone is insufficient. A successful model must follow prescribed entity trajectories, camera motion, and event timelines, while maintaining consistency over long horizons and across synchronized views. We therefore organize the evaluation into a main track of five-second clips and a challenge track of long-horizon and multi-view settings. Sections~\ref{sec:main_track} and~\ref{sec:challenge_track} report the results, and Section~\ref{sec:analysis} compares the method classes across both tracks and draws the main takeaways. The experimental setup is detailed in Section~\ref{sec:exp_setup}.

\subsection{Evaluated Methods}
\label{sec:methods}

We evaluate 11 models, grouped into three families by the control information they receive.
% (the Input column of the tables)
% The Trained column of the tables marks whether a method has been further trained on top of a video foundation model (\cmark) or uses a video foundation model as is (\xmark). 
%
% \noindent 
\textbf{(a) Camera-conditioned world models.}
EchoWM~\citep{zhang2026echowm}, SANA-WM~\citep{zhu2026sana}, LingBot-World~2.0~\citep{robbyant2026lingbotworld} and AlayaWorld~v1.1~\citep{team2026alayaworldv11} receive an RGB first frame, the text prompt, and the recorded camera trajectory. They obtain explicit camera control but no entity trajectories.
%\TODO{we didn't mention prompt here}
%
% \noindent 
\textbf{(b) Proxy-video-to-RGB models.}
These models translate an engine-rendered control representation into a realistic RGB video. Cosmos-Transfer2.5~\citep{ali2025world} receives engine depth and class-level segmentation; C2R~\citep{gomez2026coarse} receives the coarse 3D  proxy and the text; CWM~\citep{chen2026code} (fine-tuned from MiniMax-H3) receives the CWM proxy, depth, the RGB first frame, and the text; 
 LynnReal-Omni~\citep{mao2026lynnrealomninativemultimodalvideo}, which builds directly on MiniMax-H3 and extends it with native support for heterogeneous visual controls, including editable 3D renders and game recordings;
and PWM~\citep{huang2026programmableworldmodel} receives colored-OBB proxy, the RGB first frame, and the text prompt.
Because PWM is trained only on gunfight scenes, it is evaluated only on the gunfight set of Section~\ref{sec:pwm}.
%
% \noindent 
\textbf{(c) Video foundation models.}
MiniMax-H3~\citep{minimaxh3}, Seedance~2.5~\citep{seedance25} are used without any adaptation to our proxy representations. They take the colored-OBB proxy video as a reference video together with the text prompt, and RGB references as specified in each table.

\subsection{Experimental Setup}
\label{sec:exp_setup}

All evaluation samples come from recorded game-world states and are scored against the engine state; each sample is one scene observed by one camera. Every recorded state is provided in three proxy representations, Coarse 3D, CWM proxy and Colored-OBB, together with engine depth, the camera trajectory, a scene description and the state of every object, and each method receives the control representation of its native interface. Text prompts are generated automatically from the recorded scene, entities, and action timeline. All models are evaluated in their native output formats against the same engine records; the engine renders of the proxies are scored as diagnostic references. The tracks and settings are defined in Section~\ref{sec:tracks}. Methods whose native interface requires an RGB image (the world models and CWM) receive frame~0 of the MiniMax-H3 output for the same scene and camera; this frame is not a prescribed reference, but it can transmit MiniMax-H3's appearance assignments, and analyses of appearance identity account separately for effects inherited from it.

\begin{table*}[!t]
\centering
\caption{\textbf{\tocheck{Main track.}} \tocheck{Results on the Verified-FF setting (top, 40 five-second scene-camera pairs) and the Unverified-FF setting (bottom, 90 pairs); the two settings use different scenes, so best and second-best values are marked within each setting in \best{bold} and \second{underlined}.} \tocheck{Input: \inpkey{FF}{first frame checked against the recorded state (Verified-FF)}, \inpkey{UFF}{unchecked first frame, i.e., frame 0 of the MiniMax-H3 output for the same scene and camera (Unverified-FF)}, \inpkey{C}{camera pose}, \inpkey{D}{depth}, \inpkey{CS}{class-level segmentation}, \inpkey{M}{CWM proxy}, \inpkey{P}{coarse 3D proxy}, \inpkey{B}{colored-OBB proxy}.} \tocheck{All methods also receive the text prompt.} \tocheck{\ua/\da: higher/lower is better. CErr: normalized box-center error; Rot/Trans: camera rotation (degrees)/translation error; Img/Aes: imaging/aesthetic quality; Warp: warping error ($\times10^{-3}$); Smooth: motion smoothness; CLIP: CLIP-Score; ISR: Interaction Success Rate; LRA: Logic-Render Alignment; gISR/gLRA: ISR/LRA multiplied by BBox IoU per sample and then averaged, i.e., events realized by correctly placed entities.}}
\label{tab:3d_bbox}
\label{tab:dv200_firstframe}
\label{tab:dv200_nofirstframe}
\tablestyle{2.5pt}{1.12}\scriptsize
\renewcommand{\inpvocab}{FF,C,D,CS,M,P,B}
\begin{tabular}{@{}r@{\hspace{8pt}}l*{14}{c}@{}}
\toprule
\multirow{2}{*}{\textbf{Method}} & \multirow{2}{*}{\textbf{Input}} & \multicolumn{2}{c}{\textbf{Entity control}} & \multicolumn{4}{c}{\textbf{Logic / State}} & \multicolumn{3}{c}{\textbf{Camera control}} & \multicolumn{2}{c}{\textbf{Quality}} & \multicolumn{2}{c}{\textbf{Temporal}} & \textbf{Text} \\
\cmidrule(lr){3-4}\cmidrule(lr){5-8}\cmidrule(lr){9-11}\cmidrule(lr){12-13}\cmidrule(lr){14-15}\cmidrule(l){16-16}
  &  & IoU\ua & CErr\da & ISR\ua & LRA\ua & gISR\ua & gLRA\ua & Rot\da & Trans\da & CamMC\da & Img\ua & Aes\ua & Warp\da & Smooth\ua & CLIP\ua \\
\midrule
\rowcolor{black!6}\multicolumn{16}{@{}l}{\tocheck{\textit{Verified-FF (40 pairs)}}} \\
                EchoWM~\citep{zhang2026echowm} & \inp{FF,C} & 0.354 & 0.055 & 0.512 & 0.680 & 0.170 & 0.242 & \second{1.8} & 0.135 & 0.147 & 68.1 & 0.609 & 0.647 & 0.987 & 25.2 \\
        SANA-WM~\citep{zhu2026sana} & \inp{FF,C} & 0.306 & 0.070 & 0.590 & 0.569 & 0.183 & 0.179 & 2.7 & 0.145 & 0.168 & 68.6 & 0.611 & 0.871 & 0.985 & 25.1 \\
       LingBot 2.0~\citep{robbyant2026lingbotworld} & \inp{FF,C} & 0.325 & 0.064 & \best{0.727} & \second{0.804} & 0.238 & 0.258 & 2.8 & 0.100 & \second{0.132} & 70.0 & 0.615 & 1.127 & 0.977 & \second{25.4} \\
        AlayaWorld v1.1~\citep{team2026alayaworldv11} & \inp{FF,C} & 0.303 & 0.073 & 0.590 & 0.627 & 0.175 & 0.186 & 9.4 & 0.229 & 0.356 & 66.8 & 0.593 & 0.527 & 0.990 & 25.0 \\
        \arrayrulecolor{black!25}\midrule\arrayrulecolor{black}
        Cosmos 2.5~\citep{ali2025world} & \inp{D,CS} & 0.452 & 0.061 & 0.381 & 0.560 & 0.165 & 0.246 & \best{1.7} & 0.161 & 0.172 & 67.2 & 0.551 & 0.429 & \second{0.991} & 24.2 \\
        C2R~\citep{gomez2026coarse} & \inp{P} & 0.377 & 0.062 & 0.606 & 0.699 & 0.219 & 0.262 & 7.8 & 0.346 & 0.438 & \second{70.5} & 0.600 & 0.783 & 0.990 & \best{25.7} \\
        CWM~\citep{chen2026code} & \inp{M,FF,D} & 0.455 & 0.040 & 0.637 & 0.795 & 0.290 & 0.360 & 2.0 & 0.130 & 0.154 & 66.7 & \second{0.629} & \second{0.398} & \second{0.991} & 25.3 \\
        LynnReal-Omni~\citep{mao2026lynnrealomninativemultimodalvideo} & \inp{FF,B} & \best{0.518} & \best{0.037} & 0.621 & \best{0.832} & \second{0.327} & \best{0.429} & 2.1 & \best{0.074} & \best{0.101} & \best{73.8} & 0.625 & 1.258 & 0.982 & 25.2 \\
        \arrayrulecolor{black!25}\midrule\arrayrulecolor{black}
        Seedance 2.5~\citep{seedance25} & \inp{FF,B} & 0.433 & 0.043 & 0.652 & \best{0.832} & 0.280 & 0.358 & 3.6 & 0.101 & 0.152 & 66.9 & \second{0.629} & \best{0.313} & \best{0.993} & 25.3 \\
        MiniMax-H3~\citep{minimaxh3} & \inp{FF,B} & \second{0.494} & \second{0.038} & \second{0.669} & 0.748 & \best{0.330} & \second{0.372} & 3.4 & \second{0.080} & 0.133 & 70.1 & \best{0.633} & 0.444 & \second{0.991} & 25.2 \\

\midrule
\rowcolor{black!6}\multicolumn{16}{@{}l}{\tocheck{\textit{Unverified-FF (90 pairs)}}} \\
        EchoWM~\citep{zhang2026echowm} & \inp{UFF,C} & 0.255 & 0.099 & 0.669 & 0.712 & 0.172 & 0.180 & \second{3.1} & 0.165 & 0.193 & 70.2 & 0.629 & 0.823 & 0.986 & 24.8 \\
        SANA-WM~\citep{zhu2026sana} & \inp{UFF,C} & 0.240 & 0.075 & 0.564 & 0.591 & 0.145 & 0.142 & 4.7 & 0.228 & 0.275 & 69.9 & 0.636 & 0.676 & 0.986 & 24.5 \\
        %LingBot 2.0~\citep{robbyant2026lingbotworld} & \inp{UFF,C} & 0.739 & 0.811 & 0.205 & 0.110 & 3.8 & \second{0.111} & \second{0.158} & \second{72.8} & 0.633 & 1.203 & 0.973 & \second{25.1} \\
         LingBot 2.0~\citep{robbyant2026lingbotworld} & \inp{UFF,C} & 0.240 & 0.083 & 0.713 & 0.799 & 0.180 & 0.196 & 4.0 & \second{0.104} & \second{0.152} & 71.8 & 0.634 & 1.037 & 0.978 & \second{25.0} \\
        AlayaWorld v1.1~\citep{team2026alayaworldv11} & \inp{UFF,C} & 0.204 & 0.101 & 0.714 & 0.712 & 0.159 & 0.146 & 10.7 & 0.324 & 0.461 & 69.3 & 0.602 & 0.564 & \second{0.990} & 24.8 \\
       \arrayrulecolor{black!25}\midrule\arrayrulecolor{black}
        Cosmos 2.5~\citep{ali2025world} & \inp{D,CS} & 0.376 & 0.057 & 0.440 & 0.457 & 0.154 & 0.158 & 5.3 & 0.259 & 0.326 & \second{71.9} & 0.551 & 0.603 & \second{0.990} & 24.3 \\
        % \midrule
        % C2R~\citep{gomez2026coarse} & \cmark & P & 0.308 & 0.051 & 8.5 & 0.435 & 0.525 & 71.6 & 0.617 & 0.701 & 0.990 & \textbf{25.2} & 0.557 & 0.688 \\
        C2R~\citep{gomez2026coarse} & \inp{P} & 0.315 & 0.052 & 0.540 & 0.625 & 0.181 & 0.204 & 8.8 & 0.456 & 0.548 & 69.9 & 0.609 & 0.666 & \second{0.990} & \best{25.3} \\
        CWM~\citep{chen2026code} & \inp{M,UFF,D} & 0.356 & \second{0.050} & 0.741 & 0.784 & 0.269 & 0.284 & 4.7 & 0.151 & 0.208 & 67.5 & 0.637 & \second{0.384} & \second{0.990} & 24.8 \\
        LynnReal-Omni~\citep{mao2026lynnrealomninativemultimodalvideo} & \inp{B} & \best{0.457} & \best{0.036} & \best{0.809} & \best{0.852} & \best{0.379} & \best{0.395} & \best{1.4} & \best{0.084} & \best{0.101} & \best{74.8} & \best{0.670} & 1.676 & 0.976 & 24.2 \\
        \arrayrulecolor{black!25}\midrule\arrayrulecolor{black}
        Seedance 2.5~\citep{seedance25} & \inp{B} & 0.336 & 0.058 & \second{0.758} & \second{0.829} & 0.259 & 0.282 & 4.5 & 0.164 & 0.216 & 66.4 & 0.616 & \best{0.257} & \best{0.993} & 24.9 \\
        MiniMax-H3~\citep{minimaxh3} & \inp{B} & \second{0.409} & 0.055 & \second{0.758} & 0.806 & \second{0.326} & \second{0.337} & 3.6 & 0.119 & 0.168 & 71.6 & \second{0.663} & 0.407 & \second{0.990} & 24.8 \\
\bottomrule
\end{tabular}
\end{table*}

\subsection{\texorpdfstring{\tocheck{Main Track}}{Main Track}}
\label{sec:main_track}
\label{sec:main_results}
\label{sec:noff}

Table~\ref{tab:dv200_firstframe} reports ten methods in the two main-track settings.

\noindent \tocheck{\textbf{Entity Control.}
Proxy-conditioned models lead in both settings. In Verified-FF (Table~\ref{tab:dv200_firstframe}, top),} LynnReal-Omni, MiniMax-H3 and CWM reach a BBox IoU of $0.518$, $0.494$ and $0.455$, whereas the four camera-conditioned world models, which receive the camera trajectory but no entity trajectories, reach only $0.303$--$0.354$. \tocheck{In Unverified-FF (Table~\ref{tab:dv200_nofirstframe}, bottom), LynnReal-Omni and MiniMax-H3 reach $0.457$ and $0.409$, versus $0.204$--$0.255$ for the camera-conditioned world models.}

\noindent \tocheck{\textbf{Logic and State.}
LingBot~2.0 has the highest Interaction Success (ISR) in Verified-FF ($0.727$) despite the third-lowest BBox IoU ($0.325$), whereas Cosmos-Transfer2.5, which receives dense depth and segmentation, ranks fourth and third in BBox IoU but has the lowest Interaction Success (ISR) and Logic-Render Alignment (LRA) in both settings. LynnReal-Omni and Seedance~2.5 reach the highest Logic-Render Alignment (LRA) in both settings (both $0.832$ in Verified-FF; $0.852$ and $0.829$ in Unverified-FF).
Weighting the behavioral scores by entity placement changes the ordering: in Verified-FF, MiniMax-H3 achieves the highest gISR ($0.330$), followed by LynnReal-Omni ($0.327$), while LynnReal-Omni leads in gLRA ($0.429$), followed by MiniMax-H3 ($0.372$). LingBot~2.0 ranks fifth in gISR ($0.238$), showing that its leading unweighted interaction score does not translate into the strongest joint score for interaction success and spatial alignment. In Unverified-FF, LynnReal-Omni leads in both gISR ($0.379$) and gLRA ($0.395$), followed by MiniMax-H3 ($0.326$ and $0.337$).}

\noindent \tocheck{\textbf{Camera Control.}} The gap in camera control is much smaller: \tocheck{in Verified-FF,} the rotation error of most methods ($1.7$--$3.6^\circ$) is comparable to that of the pose estimator on the engine renders ($1.1$--$2.7^\circ$), and only AlayaWorld~v1.1 and C2R are clearly worse. \tocheck{LynnReal-Omni has the lowest CamMC in both settings.}

\subsection{\texorpdfstring{\tocheck{Challenge Track}}{Challenge Track}}
\label{sec:challenge_track}
\label{sec:long_horizon}
\label{sec:multiview}

\tocheck{Tables~\ref{tab:long_horizon} and~\ref{tab:multiview_crossview} report evaluation results in the long-horizon and multi-view settings of the challenge track.}

\tocheck{In the long-horizon setting (Table~\ref{tab:long_horizon}),} over the full 30\,s, BBox IoU is only $0.175$--$0.280$. The Reappearance IoU, measured on entities that return after leaving the frame, is only $0.053$--$0.131$. State Persistence, which checks whether a returning entity shows its prescribed walking or standing state, ranks the methods differently from Reappearance IoU: CWM has the lowest Reappearance IoU but a State Persistence of $0.886$. The state scores cover only 17--21 events per method from an uncalibrated judge, so these numbers indicate tendencies rather than a stable ranking. The window-chaining protocols also differ (for example, Seedance~2.5 additionally receives the last frame of the previous window), so this section measures end-to-end long-horizon preservation under each method's actual protocol, not an isolated internal memory.

\tocheck{In the multi-view setting (Table~\ref{tab:multiview_crossview}),} C2R obtains the lowest MEt3R point estimate ($0.397$), yet achieves only $32.2\%$ Appearance Compliance and $25.5\%$ Cross-view Consistency. In contrast, methods whose inputs explicitly distinguish the participants reach $82$--$94\%$ Appearance Compliance and $81$--$90\%$ Cross-view Consistency on judgeable observations.

\begin{table}[t]
\centering
\caption{\textbf{Long-horizon evaluation of controllability and memory.} Results on ten 30-second recordings, with one camera per recording and one output per method; \tocheck{best and second-best values per column in \best{bold} and \second{underlined}. Input: \inpkey{UFF}{unchecked first frame}, i.e., frame 0 of the MiniMax-H3 output for the same scene and camera, \inpkey{D}{depth}, \inpkey{M}{CWM proxy}, \inpkey{P}{coarse 3D proxy}, \inpkey{B}{colored-OBB}. \ua/\da: higher/lower is better; Rot in degrees. Re-IoU: Reappearance IoU; Re-State: State Persistence of returning entities; other metrics as in Table~\ref{tab:dv200_firstframe}.} }
\label{tab:long_horizon}
\tablestyle{4pt}{1.12}\scriptsize
\renewcommand{\inpvocab}{FF,D,M,P,B}
\begin{tabular}{@{}r@{\hspace{6pt}}l*{11}{c}@{}}
\toprule
\textbf{Method} & \textbf{Input} & \multicolumn{2}{c}{\textbf{Entity control}} & \multicolumn{4}{c}{\textbf{{Logic / State}}} & \multicolumn{2}{c}{\textbf{Memory}} & \multicolumn{3}{c}{\textbf{Camera control}} \\
\cmidrule(lr){3-4}\cmidrule(lr){5-8}\cmidrule(lr){9-10}\cmidrule(l){11-13}
 &  & IoU\ua & CErr\da & {ISR\ua} & {LRA\ua} & {gISR\ua} & {gLRA\ua} & Re-IoU\ua & Re-State\ua & Rot\da & Trans\da & CamMC\da \\
\midrule
        C2R~\citep{gomez2026coarse} & \inp{P} & \second{0.212} & \second{0.066} & {\second{0.519}} & {0.718} & {\second{0.127}} & {0.170} & \second{0.111} & 0.405 & 41.9 & 0.571 & 1.119 \\
        CWM~\citep{chen2026code} & \inp{M,UFF,D} & 0.210 & 0.070 & {0.504} & {\second{0.854}} & {0.111} & {\second{0.182}} & 0.053 & \best{0.886} & \second{30.2} & 0.381 & \second{0.793} \\
        LynnReal-Omni~\citep{mao2026lynnrealomninativemultimodalvideo} & \inp{B} & \best{0.280} & \best{0.054} & {0.515} & {\best{0.897}} & {\best{0.149}} & {\best{0.247}} & 0.103 & 0.528 & \best{22.7} & \second{0.273} & \best{0.610} \\
        \midrule
        Seedance 2.5~\citep{seedance25} & \inp{B} & 0.175 & 0.077 & {0.497} & {0.777} & {0.087} & {0.132} & \best{0.131} & 0.650 & 39.4 & \best{0.242} & 0.819 \\
        MiniMax-H3~\citep{minimaxh3} & \inp{B} & 0.206 & 0.088 & {\best{0.553}} & {0.844} & {0.121} & {0.179} & 0.059 & \second{0.833} & 31.8 & 0.351 & 0.814 \\
\bottomrule
\end{tabular}
% \begin{tabular}{@{}r@{\hspace{6pt}}l*{9}{c}@{}}
% \toprule
% \textbf{Method} & \textbf{Input} & \multicolumn{2}{c}{\textbf{Entity control}} & \multicolumn{2}{c}{\textbf{Logic / State}} & \multicolumn{2}{c}{\textbf{Memory}} & \multicolumn{3}{c}{\textbf{Camera control}} \\
% \cmidrule(lr){3-4}\cmidrule(lr){5-6}\cmidrule(lr){7-8}\cmidrule(l){9-11}
%   &  & IoU\ua & CErr\da & {ISR\ua} & {LRA\ua} & Re-IoU\ua & Re-State\ua & Rot\da & Trans\da & CamMC\da \\
% \midrule
%         C2R~\citep{gomez2026coarse} & \inp{P} & \second{0.212} & \second{0.066} & {\second{0.519}} & {0.718} & \second{0.111} & 0.405 & 41.9 & 0.571 & 1.119 \\
%         CWM~\citep{chen2026code} & \inp{M,UFF,D} & 0.210 & 0.070 & {0.504} & {\second{0.854}} & 0.053 & \best{0.886} & \second{30.2} & 0.381 & \second{0.793} \\
%         LynnReal-Omni~\citep{mao2026lynnrealomninativemultimodalvideo} & \inp{B} & \best{0.280} & \best{0.054} & {0.515} & {\best{0.897}} & 0.103 & 0.528 & \best{22.7} & \second{0.273} & \best{0.610} \\
%         \midrule
%         Seedance 2.5~\citep{seedance25} & \inp{B} & 0.175 & 0.077 & {0.497} & {0.777} & \best{0.131} & 0.650 & 39.4 & \best{0.242} & 0.819 \\
%         MiniMax-H3~\citep{minimaxh3} & \inp{B} & 0.206 & 0.088 & {\best{0.553}} & {0.844} & 0.059 & \second{0.833} & 31.8 & 0.351 & 0.814 \\
% \bottomrule
% \end{tabular}

\end{table}

\begin{table}[t]
\centering
\caption{\textbf{Multi-view evaluation of controllability and consistency.} Five methods are evaluated on 10 scenes with four synchronized views each (40 clips per method). Views are generated independently; \tocheck{best and second-best values per column in \best{bold} and \second{underlined}. Input: \inpkey{UFF}{unchecked first frame, i.e., frame 0 of the MiniMax-H3 output for the same scene and camera}, \inpkey{D}{depth}, \inpkey{M}{CWM proxy}, \inpkey{P}{coarse 3D proxy}, \inpkey{B}{colored-OBB proxy}. All methods also receive the text prompt. \ua/\da: higher/lower is better; Rot in degrees. Comp./Cons.: appearance compliance/cross-view consistency (\%); Miss: miss rate; other metrics as in Table~\ref{tab:dv200_firstframe}.} %\TODO{Explain the $^{\dagger}$ on the Camera columns, or remove it.}
}
\label{tab:multiview_crossview}
\tablestyle{3pt}{1.12}\scriptsize
\renewcommand{\inpvocab}{FF,D,M,P,B}
\begin{tabular}{@{}r@{\hspace{6pt}}l*{13}{c}@{}}
\toprule
\textbf{Method} & \textbf{Input} & \multicolumn{3}{c}{\textbf{Entity control}} & \multicolumn{4}{c}{\textbf{Logic / State}} & \multicolumn{3}{c}{\textbf{Appearance}} & \multicolumn{3}{c}{\textbf{Camera control}} \\
\cmidrule(lr){3-5}\cmidrule(lr){6-9}\cmidrule(lr){10-12}\cmidrule(l){13-15}
&  & IoU\ua & CErr\da & Miss\da & ISR\ua & LRA\ua & {gISR\ua} & {gLRA\ua} & Comp.\ua & Cons.\ua & MEt3R\da & Rot\da & Trans\da & CamMC\da \\
\midrule
        C2R~\citep{gomez2026coarse}& \inp{P} & 0.414 & 0.039 & 0.192 & 0.637 & 0.750 & {0.271} & {0.318} & 32.2 & 25.5 & \best{0.397} & 2.2 & 0.155 & 0.173 \\
        CWM~\citep{chen2026code}& \inp{M,UFF,D} & 0.439 & \second{0.037} & 0.064 & \second{0.812} & \second{0.950} & {0.345} & {0.414} & \second{92.9} & 88.1 & 0.422 & 2.0 & 0.047 & 0.071 \\
        LynnReal-Omni~\citep{mao2026lynnrealomninativemultimodalvideo}& \inp{B} & \second{0.490} & 0.038 & \best{0.047} & 0.700 & 0.900 & {\second{0.347}} & {\second{0.435}} & 87.6 & \best{89.7} & 0.448 & \best{0.4} & \best{0.023} & \best{0.026} \\
        \midrule
        Seedance 2.5~\citep{seedance25}& \inp{B} & 0.341 & 0.049 & 0.089 & \second{0.812} & \best{0.963} & {0.284} & {0.331} & 82.2 & 81.0 & 0.499 & 1.0 & 0.076 & 0.083 \\
        MiniMax-H3~\citep{minimaxh3}& \inp{B} & \best{0.524} & \best{0.029} & \second{0.049} & \best{0.825} & 0.938 & {\best{0.428}} & {\best{0.489}} & \best{94.0} & \second{88.6} & \second{0.400} & \second{0.9} & \second{0.034} & \second{0.043} \\
\bottomrule
\end{tabular}
% \begin{tabular}{@{}r@{\hspace{6pt}}l*{11}{c}@{}}
% \toprule
% \textbf{Method} & \textbf{Input} & \multicolumn{3}{c}{\textbf{Entity control}} & \multicolumn{2}{c}{\textbf{Logic / State}} & \multicolumn{3}{c}{\textbf{Appearance}} & \multicolumn{3}{c}{\textbf{Camera control}} \\
% \cmidrule(lr){3-5}\cmidrule(lr){6-7}\cmidrule(lr){8-10}\cmidrule(l){11-13}
%   &  & IoU\ua & CErr\da & Miss\da & ISR\ua & LRA\ua & Comp.\ua & Cons.\ua & MEt3R\da & Rot\da & Trans\da & CamMC\da \\
% \midrule
%         C2R~\citep{gomez2026coarse} & \inp{P} & 0.414 & 0.039 & 0.192 & 0.637 & 0.750 & 32.2 & 25.5 & \best{0.397} & 2.2 & 0.155 & 0.173 \\
%         CWM~\citep{chen2026code} & \inp{M,UFF,D} & 0.439 & \second{0.037} & 0.064 & \second{0.812} & \second{0.950} & \second{92.9} & 88.1 & 0.422 & 2.0 & 0.047 & 0.071 \\
%         LynnReal-Omni~\citep{mao2026lynnrealomninativemultimodalvideo} & \inp{B} & \second{0.490} & 0.038 & \best{0.047} & 0.700 & 0.900 & 87.6 & \best{89.7} & 0.448 & \best{0.4} & \best{0.023} & \best{0.026} \\
%         \midrule
%         Seedance 2.5~\citep{seedance25} & \inp{B} & 0.341 & 0.049 & 0.089 & \second{0.812} & \best{0.963} & 82.2 & 81.0 & 0.499 & 1.0 & 0.076 & 0.083 \\
%         MiniMax-H3~\citep{minimaxh3} & \inp{B} & \best{0.524} & \best{0.029} & \second{0.049} & \best{0.825} & 0.938 & \best{94.0} & \second{88.6} & \second{0.400} & \second{0.9} & \second{0.034} & \second{0.043} \\
% \bottomrule
% \end{tabular}
\end{table}

\subsection{\texorpdfstring{\tocheck{Discussion and Takeaways}}{Discussion and Takeaways}}
\label{sec:analysis}

\begin{figure}[t]
    \centering
    \includegraphics[width=\linewidth]{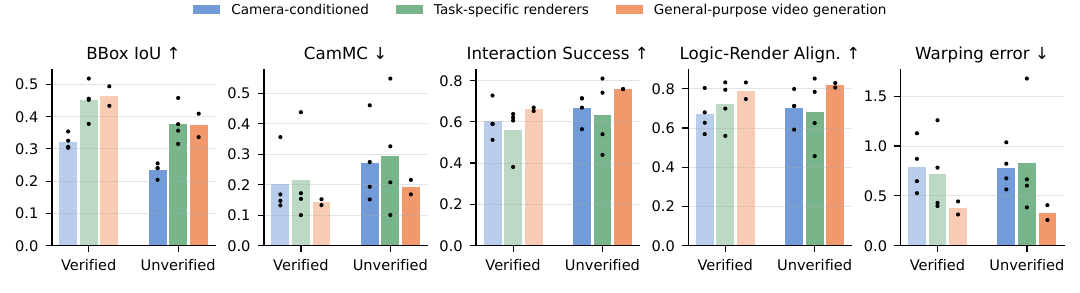}
    \caption{\textbf{\tocheck{Method classes at a glance.}} \tocheck{Main-track results per method class in Verified-FF (light) and Unverified-FF (dark). Bars show class means, dots the individual methods. Camera-conditioned: EchoWM, SANA-WM, LingBot~2.0, AlayaWorld~v1.1; task-specific renderers: Cosmos-Transfer2.5, C2R, CWM, LynnReal-Omni; general-purpose video generation: MiniMax-H3, Seedance~2.5.}}
    \label{fig:class_overview}
\end{figure}

\tocheck{We now compare the three method classes of Section~\ref{sec:methods}: camera-conditioned world models, task-specific renderers that translate a proxy into RGB video (Cosmos-Transfer2.5, C2R, CWM, LynnReal-Omni and, on the gunfight set, PWM), and general-purpose video foundation models that receive the proxy video as a reference (MiniMax-H3, Seedance~2.5). Figure~\ref{fig:class_overview} summarizes the Main-track results by class; unless stated otherwise, comparisons are paired over scenes with a sign test.} 
\tocheck{Method rankings are closely aligned across the Verified-FF and Unverified-FF settings. The Spearman rank correlations between the per-method means in the two parts of Table~\ref{tab:dv200_firstframe} are $0.98$ for BBox IoU, $0.98$ for translation error, $0.99$ for CamMC, and $0.93$ for Logic-Render Alignment. We organize the following discussion around seven takeaways.}
% These correlations indicate similar method rankings across the two disjoint scene sets, although they do not isolate the effect of first-frame conditioning, which would require paired comparisons on the same scenes under different first-frame conditions. We next distill the evaluation results into seven takeaways.}

\takeaway{T1}{\tocheck{Rendering from a proxy beats camera conditioning in following entity trajectories and event timelines.}}
\noindent\tocheck{Both proxy-conditioned classes place entities far better than the camera-conditioned world model: the class-mean BBox IoU is higher in 38 of 40 Verified-FF scenes and 87 of 89 Unverified-FF scenes (median $+0.14$, $p<10^{-8}$), with class means of $0.450$ (task-specific) and $0.464$ (general-purpose) against $0.322$ in Verified-FF, and $0.376$ and $0.373$ against $0.235$ in Unverified-FF.} In Figure~\ref{fig:dv200_firstframe_qual}, the proxy-conditioned models keep both cars in their reference boxes while one of them passes the walking protagonist (per-entity IoU $0.62$--$0.86$); EchoWM keeps the protagonist ($0.63$) but does not bring the oncoming car past the protagonist and lets the car ahead fall back toward the camera instead of keeping its distance ($0.27$ and $0.29$). \tocheck{Camera-conditioned world models do control entities implicitly, through the text prompt, the first frame and, for the protagonist, a follow camera that keeps it in view. This implicit control places the protagonist about as well as the proxy (protagonist BBox IoU $0.43$ and $0.42$ against $0.46$--$0.54$ for the proxy-conditioned classes, over third-person views) but not the other entities ($0.26$ and $0.20$ against $0.38$--$0.44$).} \tocheck{Events follow the same pattern when the proxy is paired with a strong video prior. Interaction Success checks whether each engine-recorded event reaches both its prescribed action and end state, and Logic-Render Alignment whether each segment of the prescribed timeline is visibly depicted (Figure~\ref{fig:metric_illustration}). } \tocheck{General-purpose models exceed the camera-conditioned world models in Logic-Render Alignment in both settings (26 of 32 and 57 of 77 non-tied scenes, $p<10^{-3}$) and in Interaction Success in Unverified-FF (42 of 61, $p=0.004$), and the renderers fine-tuned from MiniMax-H3 are also strong (LynnReal-Omni has the highest Logic-Render Alignment in both settings, jointly with Seedance~2.5 at the displayed precision in Verified-FF). The task-specific class as a whole, however, is not better than the camera-conditioned world models (Interaction Success in 15 of 27 and 29 of 66 scenes), because Cosmos-Transfer2.5 has the lowest Interaction Success and Logic-Render Alignment in both settings.} LingBot achieves the highest Interaction Success in Verified-FF, but correct actions and end states do not guarantee correct interaction locations, so this result should be interpreted alongside entity control results.

% \TODO{Explain why LingBot~2.0, a camera-conditioned world model without entity trajectories, reaches the highest Interaction Success in Verified-FF (0.733), although we assume that proxy-conditioned rendering models and programmable world models should realize prescribed events better. Check per-event results: which event types LingBot gets right; whether the verified first frame already shows the pre-event state so that ``no change'' passes; whether ISR is judged mainly from the protagonist, which LingBot tracks well; and whether the gap holds in Unverified-FF, where LynnReal-Omni leads.}

\takeaway{T2}{\tocheck{On camera trajectories, the two are on par: the proxy already encodes the camera.}}
\begin{figure}[t]
    \centering
    \begin{minipage}[t]{0.355\textwidth}
        \centering
        \includegraphics[width=\linewidth]{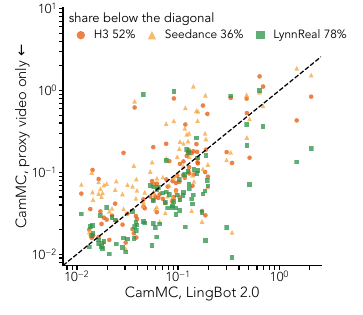}
        \caption{\textbf{\tocheck{Proxy only vs.\ camera input.}} \tocheck{Per-sample CamMC in Unverified-FF of three models without camera trajectory input against LingBot~2.0; points below the diagonal favor the proxy-only model.}}
        \label{fig:cam_scatter}
    \end{minipage}\hfill
    \begin{minipage}[t]{0.615\textwidth}
        \centering
        \includegraphics[width=\linewidth]{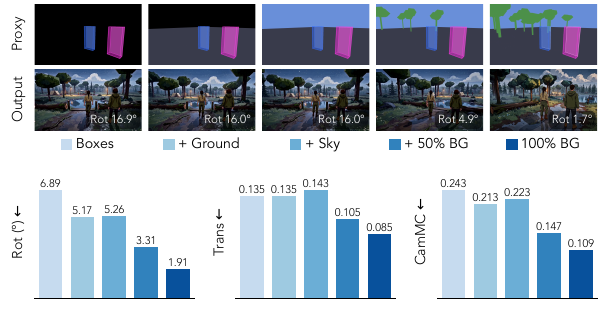}
        \caption{\textbf{\tocheck{Proxy composition and camera control.}} \tocheck{LynnReal-Omni on 20 scene-camera pairs, from entity boxes only to adding ground, sky and 50\% of the background objects. Top: scene 043 at 4.25\,s; the overlay gives the mean rotation error of that video. Bottom: means over the 20 pairs (lower is better). The 100\% BG reference uses the original prompts and first-frame setting and is not a controlled step.}}
        \label{fig:proxy_camera_control}
    \end{minipage}
\end{figure}
\noindent\tocheck{The proxy video is rendered from the prescribed camera, so it carries the camera motion implicitly. General-purpose models, which receive no camera trajectory, tie with EchoWM, SANA-WM and LingBot~2.0 in CamMC: the class mean is lower in 19 of 40 Verified-FF and 45 of 90 Unverified-FF scenes ($p=0.87$ and $p=1.0$, respectively).} \tocheck{Per sample, MiniMax-H3 does not differ significantly from EchoWM, SANA-WM and LingBot~2.0 in CamMC in Verified-FF (sign test, $p>0.2$) and has a lower translation error than EchoWM and SANA-WM in 74\% of the Unverified-FF pairs ($p<10^{-5}$), whereas LingBot~2.0 has lower camera errors than Seedance~2.5 in most pairs of both settings. MiniMax-H3, Seedance~2.5 and LynnReal-Omni all outperform AlayaWorld~v1.1 in 87--98\% of the pairs in translation error and CamMC.} \tocheck{Figure~\ref{fig:cam_scatter} compares each sample with LingBot~2.0, the camera-conditioned model with the lowest CamMC in Unverified-FF: MiniMax-H3 has the lower camera error in 52\% of the samples, Seedance~2.5 in 36\%, and LynnReal-Omni in 78\%.}

\noindent\tocheck{The composition of the proxy also matters for camera control.} \tocheck{Figure~\ref{fig:proxy_camera_control} examines how proxy components affect camera control across 20 scene-camera pairs. Starting from entity boxes, adding ground reduces CamMC from $0.243$ to $0.213$, while adding sky provides no further improvement ($0.223$). Incorporating 50\% of the background objects produces larger gains, reducing rotation error from $6.89^\circ$ to $3.31^\circ$, translation error from $0.135$ to $0.105$, and CamMC from $0.243$ to $0.147$ relative to boxes-only proxies. The illustrated scene exhibits the same pattern: mean rotation error remains near $16^\circ$ after adding ground and sky, but falls to $4.89^\circ$ when background objects are included. These results suggest that explicit background structure supplies useful spatial cues for reproducing prescribed camera motion. The full-background reference achieves still lower errors, with a rotation error of $1.91^\circ$ and CamMC of $0.109$. However, this setting also differs in prompts, first-frame inputs, and some entity-box sets, so its additional gains cannot be attributed solely to increased background coverage.}

% \takeaway{T3}{\tocheck{Off-the-shelf video foundation models are already competent renderers of short clips; fine-tuning buys control at the cost of stability.}}
% \noindent\tocheck{ On the short, single-camera clips of the main track,
% against the task-specific renderers, the general-purpose models tie in BBox IoU (23 of 40 and 43 of 89 scenes) but have lower camera errors (CamMC in 30 of 40 and 76 of 90 scenes, $p\le 0.002$), higher Interaction Success and Logic--Render Alignment in Unverified-FF (54 of 68 and 62 of 80 scenes, $p\approx10^{-6}$), and the lowest warping error (class mean $0.38$ and $0.33$ against $0.72$ and $0.83$).} \tocheck{
% CWM and LynnReal-Omni both build on MiniMax-H3, which allows paired comparisons with their general-purpose base. CWM has a lower BBox IoU than MiniMax-H3 in 71\% of the Unverified-FF pairs ($p=10^{-4}$) and a lower imaging quality in about 90\% of the pairs in both settings, but a lower warping error in about 80\%. LynnReal-Omni has a higher BBox IoU in 62\% of the Unverified-FF pairs ($p=0.03$) and a higher Logic--Render Alignment in 81\% of the non-tied Verified-FF pairs ($p=0.02$), but a higher warping error in every pair of both settings. Lower warping error can also reflect less motion; still, neither fine-tuned model improves on MiniMax-H3 across the board.}

\takeaway{T3}{\tocheck{General-purpose video generation models already render short clips well.}}
\noindent\tocheck{On the short, single-camera clips of the main track, general-purpose models are competitive with task-specific renderers across several metrics. They achieve comparable BBox IoU (22 of 40 and 43 of 89 scenes), lower camera errors (CamMC in 29 of 40 and 76 of 90 scenes, $p\le 0.007$), and higher Interaction Success and Logic-Render Alignment in Unverified-FF (53 of 67 and 61 of 79 scenes, $p\approx10^{-6}$). They also achieve the lowest warping error, with class means of $0.38$ and $0.33$, compared with $0.72$ and $0.83$ for task-specific renderers.}

\takeaway{T4}{\tocheck{Generic fine-tuning gives mixed gains; domain-specific events still need dedicated training.}}
\begin{table}[t]
\centering
\caption{\textbf{Comparison with a domain-specific model on armed-confrontation scenes.} The 20-scene set includes first-frame conditioning and is designed for comparison with PWM, which is trained only on gunfight scenes (third-person; an armed confrontation in which one person is shot and collapses). Metrics use native model outputs; \tocheck{best and second-best values are \best{bold} and \second{underlined}}. \tocheck{Input: \inpkey{FF}{first frame}, \inpkey{D}{depth}, \inpkey{M}{CWM proxy}, \inpkey{P}{coarse 3D proxy}, \inpkey{B}{colored-OBB proxy}.} \tocheck{All methods also receive the text prompt.} CWM = Code World Model. \tocheck{Metrics as in Table~\ref{tab:dv200_firstframe}.}}
\label{tab:neon_death}
\tablestyle{3pt}{1.12}\scriptsize
\renewcommand{\inpvocab}{FF,D,M,P,B}
\begin{tabular}{@{}ll*{14}{c}@{}}
\toprule
\multirow{2}{*}{\textbf{Method}} & \multirow{2}{*}{\textbf{Input}} & \multicolumn{2}{c}{\textbf{Entity control}} & \multicolumn{4}{c}{\textbf{Logic / State}} & \multicolumn{3}{c}{\textbf{Camera control}} & \multicolumn{2}{c}{\textbf{Quality}} & \multicolumn{2}{c}{\textbf{Temporal}} & \textbf{Text} \\
\cmidrule(lr){3-4}\cmidrule(lr){5-8}\cmidrule(lr){9-11}\cmidrule(lr){12-13}\cmidrule(lr){14-15}\cmidrule(l){16-16}
&  & IoU\ua & CErr\da & ISR\ua & LRA\ua & {gISR\ua} & {gLRA\ua} & Rot\da & Trans\da & CamMC\da & Img\ua & Aes\ua & Warp\da & Smooth\ua & CLIP\ua \\
\midrule
        C2R~\citep{gomez2026coarse}& \inp{P} & 0.437 & 0.041 & 0.450 & 0.725 & {0.191} & {0.330} & 2.7 & 1.008 & 1.043 & 69.6 & 0.544 & 0.301 & \second{0.994} & 24.3 \\
        CWM~\citep{chen2026code}& \inp{M,FF,D} & 0.413 & 0.039 & \second{0.650} & \second{0.900} & {0.287} & {0.381} & \best{0.7} & 0.740 & 0.744 & 72.6 & \second{0.562} & \second{0.157} & 0.992 & \second{25.6} \\
        PWM~\cite{huang2026programmableworldmodel}& \inp{FF,B} & \best{0.602} & \best{0.021} & \best{0.850} & 0.875 & {\best{0.523}} & {\second{0.529}} & \second{1.1} & \best{0.348} & \best{0.359} & 71.0 & \best{0.579} & 0.182 & 0.993 & \best{26.0} \\
        LynnReal-Omni~\citep{mao2026lynnrealomninativemultimodalvideo}& \inp{FF,B} & \second{0.552} & \second{0.033} & 0.625 & \best{0.975} & {\second{0.340}} & {\best{0.535}} & 1.3 & \second{0.563} & \second{0.584} & \best{75.0} & 0.527 & 0.613 & 0.988 & 25.2 \\
        \midrule
        MiniMax-H3~\citep{minimaxh3}& \inp{FF,B} & 0.526 & 0.035 & 0.600 & \second{0.900} & {0.302} & {0.475} & 5.1 & 0.770 & 0.830 & \second{73.9} & 0.556 & \best{0.107} & \best{0.995} & 25.4 \\
\bottomrule
\end{tabular}
\end{table}

\label{sec:pwm}
% Finally, we compare the general pipelines with a model trained for a specific domain, on events of that domain. PWM is trained on limited scenes, so we built a dedicated set of 20 gunfight scenes in which one person is shot and collapses on a prescribed schedule (Table~\ref{tab:neon_death}). \tocheck{The first frame is generated by MiniMax-H3~\citep{minimaxh3}.}
% PWM obtains a BBox IoU of $0.602$ and an Interaction Success Rate of $0.850$. MiniMax-H3 reaches a BBox IoU of $0.526$ but an Interaction Success Rate of only $0.600$: models that place the person roughly correctly can still differ clearly in whether the event is executed. 
\noindent\tocheck{Fine-tuning does not, however, yield uniform improvements. CWM and LynnReal-Omni both build on MiniMax-H3, enabling paired comparisons with their general-purpose base. CWM has lower BBox IoU in 71\% of the Unverified-FF pairs ($p=10^{-4}$) and lower imaging quality in approximately 90\% of the pairs in both settings, but lower warping error in approximately 80\%. LynnReal-Omni achieves higher BBox IoU in 62\% of the Unverified-FF pairs ($p=0.03$) and higher Logic-Render Alignment in 80\% of the non-tied Verified-FF pairs ($p=0.035$), but higher warping error in every pair in both settings. Because lower warping error can also reflect reduced motion, it should not be interpreted as an unqualified improvement in temporal consistency. These comparisons show that the benefits of fine-tuning depend on the model and the control objective.}

\begin{figure}[t]
    \centering
    \includegraphics[width=\linewidth]{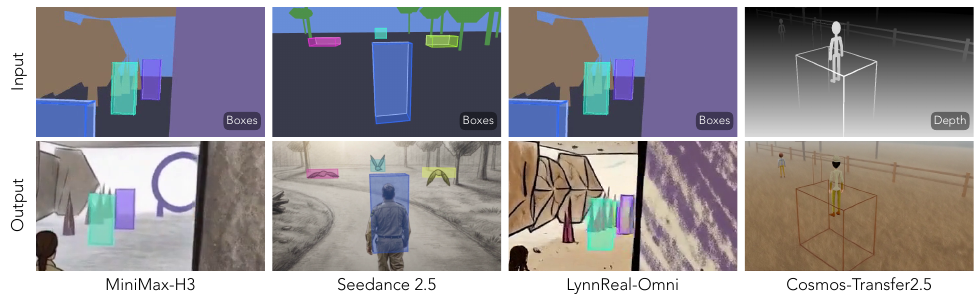}
    \caption{\textbf{Proxy leakage.} The top row shows proxy inputs, and the bottom row shows corresponding generated frames. In these representative failure cases, colored bounding boxes or wireframe outlines from the proxies remain visible in the outputs, revealing unintended leakage of control representations into the rendered appearance.}
    % \vspace{-4mm}
    \label{fig:proxy_leakage}
\end{figure}

\tocheck{Despite their strong performance on short clips, general-purpose models can still struggle to execute prescribed events reliably. We illustrate this limitation on a dedicated evaluation set of 20 gunfight scenes in which one person is shot and collapses according to a prescribed schedule (Table~\ref{tab:neon_death}). We compare the general-purpose pipelines with PWM, a domain-specific model trained on a limited set of scenes. The first frames are generated by MiniMax-H3~\citep{minimaxh3}. MiniMax-H3 achieves a BBox IoU of $0.526$ and an Interaction Success Rate of $0.600$, while PWM reaches $0.602$ and $0.850$, respectively. The gap in Interaction Success Rate highlights a limitation of general-purpose rendering: reasonable spatial alignment does not guarantee reliable execution of a prescribed event. Furthermore, as shown in Figure~\ref{fig:proxy_leakage}, outputs from general-purpose models sometimes exhibit proxy leakage, whereas such artifacts are rarely observed in PWM's outputs.}

\takeaway{T5}{\tocheck{The next step is finding the best proxy.}}
% \noindent\tocheck{ The proxy decides what the renderer is told, and its form matters little for placement.} As shown in Table~\ref{tab:h3_representations}, for a fixed generator, first frame and seed, the choice of proxy changes layout (BBox IoU 0.506–0.550),\TODO{Stale range from the previous Table~\ref{tab:h3_representations}; the current Verified-FF subset gives BBox IoU 0.473--0.481. Update the numbers and recheck this conclusion.}\TODO{State which set and how many clips Table~\ref{tab:h3_representations} uses (it is not the main-track set), and whether the ISR/LRA values are comparable with Table~\ref{tab:dv200_firstframe}.} 
\noindent\tocheck{The proxy decides what the renderer is told, but its form matters little for placement. Table~\ref{tab:h3_representations} compares three proxies on ten Verified-FF pairs (nine follow, one first-person): the coarse-3D-proxy, CWM-proxy  and colored-OBB rows are generated with the same first frames, seed and generation settings. All metrics follow the protocol of Table~\ref{tab:dv200_firstframe} but cover only this subset. The layout barely changes with the proxy (BBox IoU $0.503$--$0.514$), and camera and quality metrics differ little. The event metrics differ more, but only through a few scenes: the colored-OBB proxy has the highest Interaction Success ($0.733$ against $0.625$--$0.683$; higher than the coarse 3D proxy in three scenes and lower in none, and higher than the CWM proxy in two scenes and lower in one), and the coarse 3D proxy has the highest Logic--Render Alignment ($0.802$; higher than the colored-OBB proxy in four scenes and lower in one); with ten pairs, none of these differences is significant (sign test, $p\ge0.25$).} 
From our visualizations, we observe that, compared with coarse 3D and CWM proxies, colored-OBB proxies generate more natural human motions, with richer joint articulation and more refined strides. Therefore, we use the colored-OBB proxy as the representation input to MiniMax-H3 in our evaluation.
% In our examples, box proxies also yield more articulated strides than the coarse 3D and CWM proxies.
% while in our examples box proxies yield more articulated strides than coarse 3D proxy and
% mixed proxies.
% We therefore use the bounding-box proxy as the representation input to Minimax-H3 in the evaluation.

% \begin{figure}[H]
%   \centering
%   % \includegraphics[width=\linewidth]{figures/teaser.pdf}
%   \includegraphics[width=\linewidth]{figures/proxy.pdf}
%   \caption{\textbf{Overview of representation }
%   }
%   \label{fig:different-proxy}
%   % \vspace{-2mm}
% \end{figure}

\tocheck{However, the proxy leaves a key ambiguity unresolved: which appearance belongs to which entity. A box proxy fixes where each entity is and how it moves, but not what it looks like; which described appearance belongs to which box must be inferred from the text.} Without a visual reference, the model must infer from text which appearance description belongs to which person. As a result, even when the overall scene and actions appear plausible, identity--appearance associations can be incorrect. In the supermarket example in Figure~\ref{qual_nofirstframe_identity}a \tocheck{(Appendix~\ref{app:qual})}, which contains a courier, a student, and a tourist, MiniMax-H3 swaps the student's and tourist's outfits throughout the clip. Seedance~2.5 assigns the courier's outfit to the student's position while leaving the courier's box empty, whereas LynnReal-Omni preserves the intended appearance assignment for all three characters. In a manual audit of 62 multi-person scenes, MiniMax-H3 swaps the appearances of two people in 12 cases. These examples suggest that visually plausible multi-person generation can still exhibit errors in maintaining entity identity and associated appearance attributes. 

% \TODO{Check whether CWM's representation (CWM proxy with semantic identities, depth and the RGB first frame) avoids this ambiguity. In the manual audit, CWM swaps appearances in 10 of the 62 scenes, all inherited from MiniMax-H3's frame~0 in Unverified-FF; the CWM proxy itself may not disambiguate people of the same class. Compare with LynnReal-Omni (2 swaps) and quantify per representation.}

\tocheck{The same ambiguity affects orientation.} Without an image that fixes how each entity looks from the camera, models sometimes render an entity facing the opposite way. In the horseback scene of Figure~\ref{qual_nofirstframe_identity}b, the horse on the right faces away from the camera: LynnReal-Omni and MiniMax-H3 show its rear, whereas Seedance~2.5 renders it facing the camera. The entity is placed in its box, but its front--back direction is not controlled reliably. \tocheck{The mixed representation in Figure~\ref{fig:handoff_all} (Appendix~\ref{app:qual}) likewise omits limbs, whereas the coarse 3D proxy keeps them. A proxy that encodes identity, facing direction and articulation while remaining cheap to render is therefore a direct lever for faithful rendering, and the benchmark can compare candidate proxies under a fixed generator as in Table~\ref{tab:h3_representations}.}

% \TODO{SAY box is better than others in several cases. MAYBE need put a few samples to compare}

\begin{table}[h]
\centering
\caption{\textbf{MiniMax-H3 under different proxy representations on a Verified-FF subset.} \tocheck{Input: \inpkey{FF}{first frame}, \inpkey{M}{CWM proxy}, \inpkey{P}{coarse 3D proxy}, \inpkey{B}{colored-OBB proxy}.} \tocheck{Metrics as in Table~\ref{tab:dv200_firstframe}.}}
\label{tab:h3_representations}
\tablestyle{3pt}{1.12}\scriptsize
\renewcommand{\inpvocab}{FF,M,P,B}
\begin{tabular}{@{}ll*{14}{c}@{}}
\toprule
\multirow{2}{*}{\textbf{Method}} & \multirow{2}{*}{\textbf{Input}} & \multicolumn{2}{c}{\textbf{Entity control}} & \multicolumn{4}{c}{\textbf{Logic / State}} & \multicolumn{3}{c}{\textbf{Camera control}} & \multicolumn{2}{c}{\textbf{Quality}} & \multicolumn{2}{c}{\textbf{Temporal}} & \textbf{Text} \\
\cmidrule(lr){3-4}\cmidrule(lr){5-8}\cmidrule(lr){9-11}\cmidrule(lr){12-13}\cmidrule(lr){14-15}\cmidrule(l){16-16}
&  & IoU\ua & CErr\da & ISR\ua & LRA\ua & {gISR\ua} & {gLRA\ua} & Rot\da & Trans\da & CamMC\da & Img\ua & Aes\ua & Warp\da & Smooth\ua & CLIP\ua \\
\midrule
        MiniMax-H3~\citep{minimaxh3}& \inp{FF,P} & \best{0.514} & \best{0.030} & 0.625 & \best{0.802} & 0.304 & \best{0.391} & \second{4.8} & \second{0.072} & \second{0.156} & 70.1 & \best{0.655} & \best{0.570} & \best{0.987} & \best{25.1} \\
        MiniMax-H3~\citep{minimaxh3}& \inp{FF,M} & \second{0.507} & \second{0.031} & \second{0.683} & 0.665 & \second{0.338} & 0.333 & \best{4.5} & 0.079 & \second{0.156} & \best{70.7} & 0.650 & \second{0.573} & \best{0.987} & \best{25.1} \\
        MiniMax-H3~\citep{minimaxh3}& \inp{FF,B} & 0.503 & 0.033 & \best{0.733} & \second{0.690} & \best{0.347} & \second{0.339} & 5.0 & \best{0.065} & \best{0.155} & \best{70.7} & \second{0.652} & 0.574 & \best{0.987} & \best{25.1} \\
\bottomrule
\end{tabular}
% \begin{tabular}{@{}ll*{12}{c}@{}}
% \toprule
% \multirow{2}{*}{\textbf{Method}} & \multirow{2}{*}{\textbf{Input}} & \multicolumn{2}{c}{\textbf{Logic / State}} & \multicolumn{2}{c}{\textbf{Entity control}} & \multicolumn{3}{c}{\textbf{Camera control}} & \multicolumn{2}{c}{\textbf{Quality}} & \multicolumn{2}{c}{\textbf{Temporal}} & \textbf{Text} \\
% \cmidrule(lr){3-4}\cmidrule(lr){5-6}\cmidrule(lr){7-9}\cmidrule(lr){10-11}\cmidrule(lr){12-13}\cmidrule(l){14-14}
%  & & ISR\ua & LRA\ua & IoU\ua & CErr\da & Rot\da & Trans\da & CamMC\da & Img\ua & Aes\ua & Warp\da & Smooth\ua & CLIP\ua \\
% \midrule
%         MiniMax-H3~\citep{minimaxh3} & \inp{FF,P} & 0.948 & 0.924 & 0.550 & 0.019 & 5.94 & 0.159 & 0.257 & 76.7 & 0.541 & 0.256 & 0.990 & 26.6 \\
%         MiniMax-H3~\citep{minimaxh3} & \inp{FF,M} & 0.984 & 0.889 & 0.506 & 0.024 & 5.08 & 0.126 & 0.213 & 76.8 & 0.541 & 0.261 & 0.990 & 26.6 \\
%         MiniMax-H3~\citep{minimaxh3} & \inp{FF,B} & 0.969 & 0.864 & 0.537 & 0.023 & 6.84 & 0.155 & 0.263 & 76.7 & 0.539 & 0.263 & 0.990 & 26.7 \\
%         MiniMax-H3~\citep{minimaxh3} & \inp{FF,AB} & 0.938 & 0.888 & 0.536 & 0.022 & 6.21 & 0.138 & 0.247 & 76.8 & 0.542 & 0.260 & 0.990 & 26.8 \\
% \bottomrule
% \end{tabular}
\end{table}

\takeaway{T6}{\tocheck{Open problems remain in the challenge track: long horizons, re-entering entities and multi-view identity.}}
\begin{figure}[t]
\centering
\begin{minipage}[t]{0.355\textwidth}
\centering
\includegraphics[width=\linewidth,trim=0 21 467 0,clip]{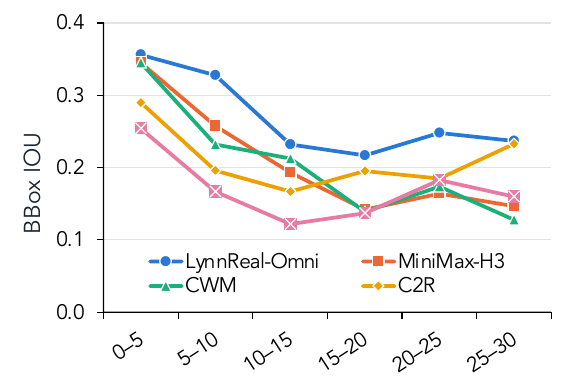}
\caption{\textbf{\tocheck{BBox IoU over the 30-s recordings.}} Averaged in 5-s windows over the ten recordings and all entities with a reference box: every method is best in the first five seconds and loses the prescribed layout as the sequence continues (the visible entities differ between windows, so the curves show a trend).}

\label{fig:long_horizon_iou}
\end{minipage}\hfill
\begin{minipage}[t]{0.615\textwidth}
\centering
\includegraphics[width=\linewidth,trim=276 21 0 0,clip]{figures/longclip.pdf}
\caption{\textbf{\tocheck{A returning pedestrian.}} Park scene, first-person camera, the recording with the largest drop: the dashed box is the engine reference of a pedestrian who leaves the view at 3.9\,s and returns at 17.25\,s. All methods localize it before it leaves (IoU about $0.6$); after the return, the IoU is $0.34$ (CWM), $0.30$ (LynnReal-Omni), $0.17$ (Seedance~2.5) and $0.01$ (MiniMax-H3).}
\label{fig:long_horizon_case}
\end{minipage}
\end{figure}

\noindent\tocheck{No class solves the challenge track: over 30\,s, BBox IoU is $0.175$--$0.280$ and returning entities reach a Reappearance IoU of only $0.053$--$0.131$ (Table~\ref{tab:long_horizon}, Figures~\ref{fig:long_horizon_iou} and~\ref{fig:long_horizon_case}). \tocheck{The long-horizon BBox IoU is} below the $0.32$--$0.52$ of the same five methods on the five-second sets (different scenes), and 42 of the 50 method--recording results stay below $0.3$. Figure~\ref{fig:long_horizon_iou} shows how this arises: every method is best in the first five seconds ($0.25$--$0.36$) and falls to $0.12$--$0.25$ after ten seconds, so the prescribed layout is followed at the start of a sequence but increasingly lost as it continues. In Figure~\ref{fig:long_horizon_case}, every method localizes a pedestrian before it leaves (IoU about $0.6$); after it re-enters some 13\,s later, the post-return IoU is $0.34$ (CWM), $0.30$ (LynnReal-Omni), $0.17$ (Seedance~2.5) and $0.01$ (MiniMax-H3). Following an entity within a local window does not imply restoring its position after a long absence across windows. \tocheck{Multi-view identity is the second open problem. The four synchronized views of a scene are generated independently, so each player must be re-created from its description in every view. Per-view control remains good (BBox IoU $0.34$--$0.52$, CamMC $0.026$--$0.173$; Table~\ref{tab:multiview_crossview}), but identity depends on whether the input binds each description to one person. Methods that receive identity-colored boxes (LynnReal-Omni, Seedance~2.5, MiniMax-H3) or a first frame (CWM) reach $82$--$94\%$ Appearance Compliance and $81$--$90\%$ Cross-view Consistency, whereas C2R, whose grey coarse 3D proxy draws all players alike and which receives no first frame, reaches $32\%$ and $26\%$. Even the best methods render a differently dressed player in $10$--$19\%$ of the judged view pairs, and these scores cover only observations the judge could assess.} \tocheck{State Persistence ranks the methods differently from Reappearance IoU (Section~\ref{sec:challenge_track}):} The two metrics capture different aspects of long-term consistency. Domain-specific events still require dedicated training: on the gunfight set, general models reach an Interaction Success of $0.450$--$0.650$ against $0.850$ for PWM. Without a verified first frame, appearance descriptions can be bound to the wrong person (T5). Together, these results suggest developing rendering models on top of strong video foundation models, targeting event fidelity, long-horizon memory and identity binding.}

\takeaway{T7}{\tocheck{The benchmark reveals what existing metrics miss, but has limitations.}}
\noindent\tocheck{ The task metrics capture what placement and perceptual metrics miss. Interaction Success and Logic-Render Alignment are largely independent of entity placement. Within a scene, the Spearman correlation between BBox IoU and Interaction Success across methods averages $0.05$ in Verified-FF and $0.06$ in Unverified-FF (and $0.14$ and $0.12$ for Logic-Render Alignment). Interaction Success and Logic-Render Alignment also differ from each other: in Figure~\ref{fig:metric_illustration}, LingBot~2.0 depicts all three prescribed actions (Logic-Render Alignment $1.00$) but misses the end state of the first event (Interaction Success $0.67$). Logic-Render Alignment responds to the prescription rather than to the video alone: when the judge receives a mismatched timeline, the per-method mean drops from $0.46$--$0.85$ to $0.07$--$0.20$, and the matched timeline scores higher in 68--97\% of the samples.} \tocheck{
Across methods, the mean CLIP-Score varies by only $4.5$--$5.7\%$ of its average and motion smoothness by approximately $1.6\%$ in both settings, compared with $53$--$79\%$ for BBox IoU and $38$--$58\%$ for Interaction Success and Logic-Render Alignment. In Unverified-FF, LynnReal-Omni has the lowest CLIP-Score ($24.2$) while leading all entity, camera and logic metrics.} \tocheck{The multi-view setting shows the same for geometric metrics:} The results reveal a clear gap between geometric agreement and participant consistency across independently generated views. Thus, a model can generate views that are geometrically coherent while failing to preserve the same participant appearance across viewpoints. Multi-view evaluation should therefore measure both geometric and identity/appearance consistency, since geometric agreement alone is not a reliable proxy for participant consistency. \tocheck{The benchmark also has limitations. Interaction Success, Logic-Render Alignment and the appearance metrics rely on a VLM judge that has not yet been calibrated against human annotation; camera metrics inherit the error of the pose estimator, which reaches $1.1$--$5.4^\circ$ ($1.1$--$2.7^\circ$ in Verified-FF) even on the engine renders; the Unverified-FF set was curated with MiniMax-H3 outputs and all first frames are generated by MiniMax-H3, which may favor it; and the challenge-track sets are small (ten recordings or scenes each).} 

% \TODO{Add a human--judge agreement study on a subset of events and segments.}

%%%%%%%%%

\section{Conclusion}

We introduced \textbf{PROWBench}, a benchmark and extensible framework that grounds the evaluation of programmable world models in replayable world states and interaction records. Synchronized visual proxies enable controlled comparisons across representations, viewpoints, and time, while Logic-Render Alignment and Interaction Success Rate complement geometric and perceptual metrics by measuring whether engine-defined rules and events are visibly realized.

Our experiments demonstrate the value of explicit entity-level control: proxy-conditioned models adhere to prescribed trajectories more closely than models receiving camera control alone, and can generate coherent scenes even without an RGB first-frame reference. At the same time, visual plausibility does not guarantee faithful world realization. Appearance assignment, identity binding, and facing direction remain unreliable without a visual reference; entity control and reappearance degrade over longer horizons; and geometric agreement across independently generated views does not ensure consistent participant identity or appearance.

These findings position programmable-world rendering as a problem of preserving executable state across entities, events, time, and viewpoints, rather than optimizing visual quality alone. PROWBench provides a reproducible foundation for measuring progress toward this goal. Reliable generative world renderers will require stronger entity--appearance binding, event-aware generation, long-horizon state memory, and joint multi-view consistency.

%%%%%%%%%%%

\bibliographystyle{abbrv}
\bibliography{references}

\newpage

\beginappendix
\suppressfloats[t]
\counterwithin{figure}{section}
\counterwithin{table}{section}

\section{\texorpdfstring{\tocheck{Benchmark Data}}{Benchmark Data}}
\label{app:data}

\tocheck{This section reports dataset statistics and describes how the first frames and text prompts of the benchmark are constructed.}

\subsection{Data Statistics}
\label{sec:data_statistics}

\begin{figure*}[p]
    \centering
    \includegraphics[width=\textwidth]{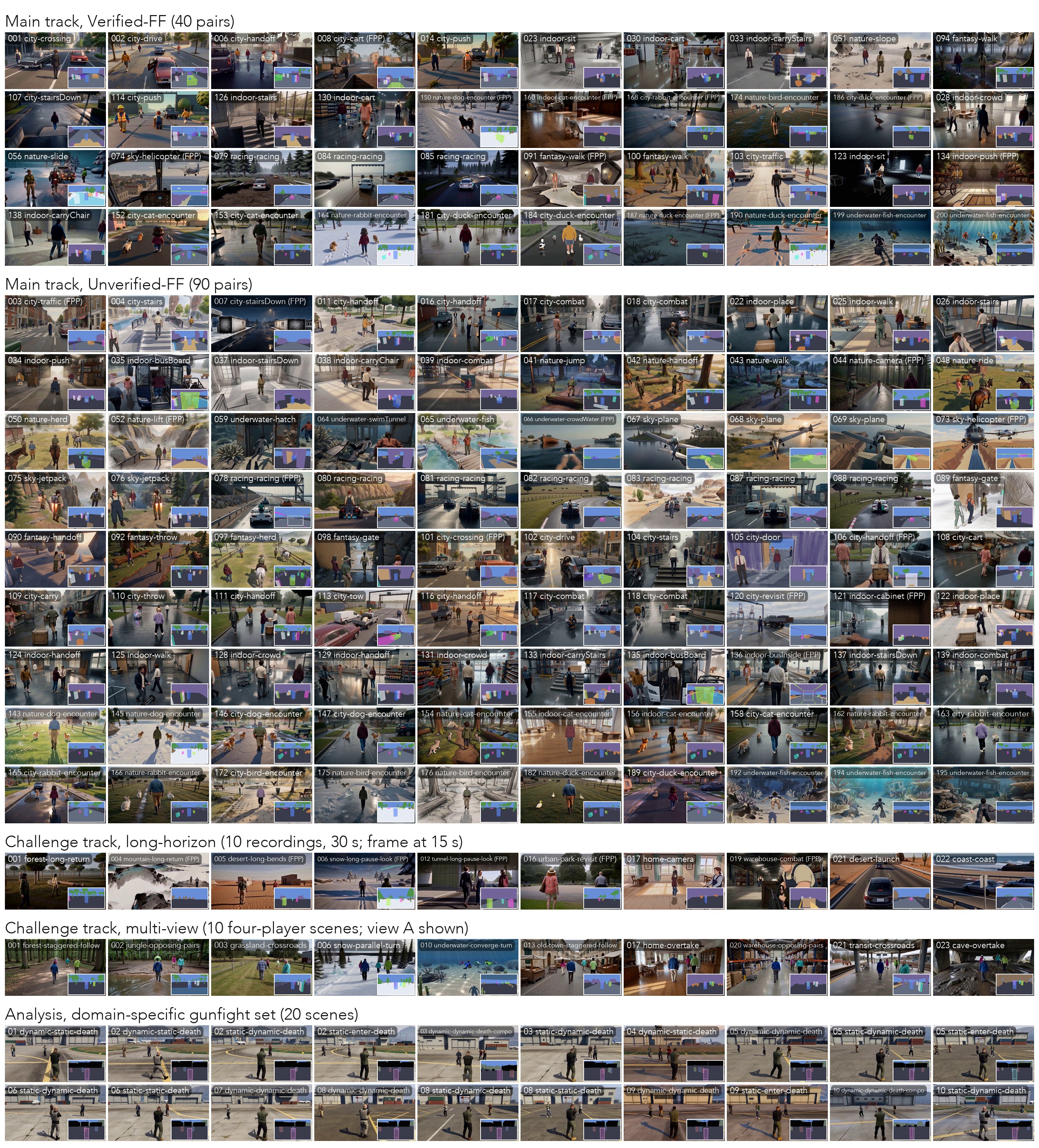}
    \caption{\textbf{\tocheck{Gallery of all 170 PROWBench cases.}} \tocheck{Each tile shows the MiniMax-H3 output at the middle of the clip (at 15\,s for the 30-second long-horizon recordings); the inset shows the colored-OBB proxy input at the same instant. MiniMax-H3 is used only because it has outputs for every collection; its frames are generated results, not ground truth. Cases are grouped by collection: the Verified-FF (40) and Unverified-FF (90) settings of the main track, the long-horizon (10) and multi-view (10, view A shown) settings of the challenge track, and the domain-specific gunfight set (20). FPP marks first-person views.}}
    \label{fig:gallery}
\end{figure*}

\begin{figure}[t]
    \centering
    \includegraphics[width=\linewidth]{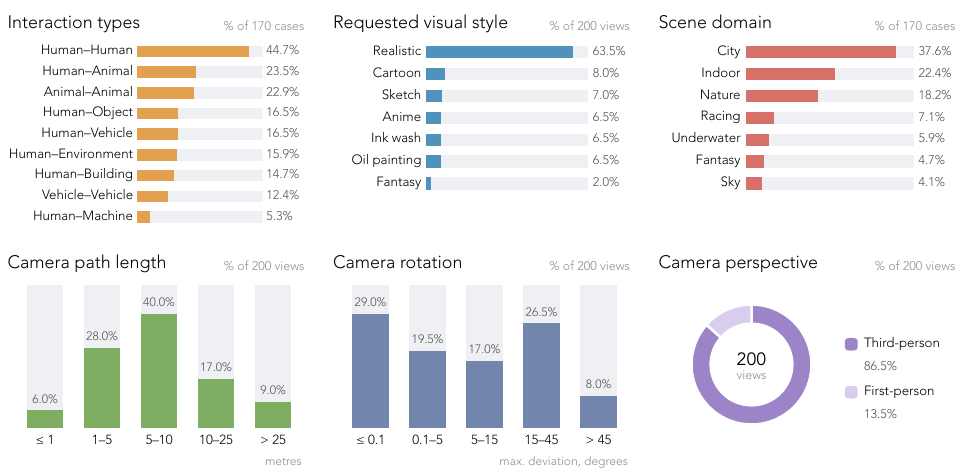}
    \caption{\textbf{Dataset composition of PROWBench.} 170
  evaluation cases, 200 selected camera views, and 600
  proxy videos. \tocheck{The panels show interaction categories, requested visual styles, scene domains, camera path length, maximum angular deviation from the initial camera orientation, and camera perspectives; each panel states its denominator (170 episodes or 200 selected views).} Interaction labels
  follow a unified task-level annotation policy and
  may overlap within a case. Camera statistics cover each full
  source clip, including both five-second and 30-
  second sequences. Each selected view provides three
  synchronized proxy representations derived from the
  same source world-state records.}
    \label{fig:diversity}
\end{figure}

\tocheck{Figure~\ref{fig:gallery} shows all 170 cases.} PROWBench comprises 170 evaluation cases identified by unique episode IDs, 200 selected camera views, and 600 proxy input videos across four collections: diverse scenarios, multi-participant trajectories, event-driven state transitions, and long-horizon trajectories. Each selected view provides three synchronized representations: coarse 3D, CWM proxy, and Colored-OBB. All source proxy inputs are rendered at 30\,FPS: 570 videos are five seconds long with 150 frames each, and 30 videos are 30 seconds long with 900 frames each. There are 540 videos at $1280 \times 720$ and 60 event-driven videos at $832 \times 480$. This yields 37,500 view-specific camera-frame observations and 112,500 proxy video frames, totaling 62.5 minutes of aggregate proxy-video duration across the three representations.

The diverse-scenario collection contains 130 episodes, divided between the \tocheck{40-case Verified-FF and 90-case Unverified-FF settings of the main track}. Each episode contributes one selected view and three proxy videos, yielding 390 videos. This collection spans seven scene domains and, under the common interaction policy below, nine non-exclusive interaction categories with 227 episode--tag assignments.

The multi-participant collection, Multi4, contains ten episodes with four participants following distinct scripted trajectories each. It covers six trajectory patterns: staggered following, opposing movement, crossing, convergence with turning, overtaking, and parallel turns. Four synchronized third-person following views are selected per episode, yielding 40 views and 120 proxy videos.

The event-driven collection, Gunfight, contains 20 cases derived from ten source recordings. Each case provides one third-person view and three proxy videos, contributing 20 views and 60 videos. These cases vary camera and actor motion around a recorded actor-death event. Episode IDs therefore identify benchmark cases and need not correspond to independent source recordings.

The long-horizon collection contains ten continuous 30-second episodes covering traversal, landmark revisitation, camera look-around, combat, and racing. One view is selected per episode: six first-person and four third-person views, yielding ten views and 30 proxy videos. These cases support the long-horizon evaluation in Section~\ref{sec:long_horizon} and are included in the dataset totals above.

Scene-domain labels combine 130 native annotations with 40 source-reviewed annotations. The resulting counts are city (64), indoor (38), nature (31), racing (12), underwater (10), fantasy (8), and sky (7); city includes airport aprons and industrial outdoor settings. We review interaction labels across all 170 episodes under a common task-level policy, using source descriptions, entity roles, recorded events and trajectories, and sampled proxy frames. Labels cover the authored focal task and its participants, including explicit relative group motion; incidental background co-occurrence is excluded. Weapons, handcarts and furniture count as objects; fixed architectural access counts as human--building interaction; landmark observation and revisitation count as environment exploration. Original labels and per-tag evidence are retained. This AI-assisted review yields 293 episode--tag assignments across nine non-exclusive categories. These are source-defined task labels, including description-supported weapon use in Gunfight, and do not imply independently verified physical contact or hit causality.

Across the four collections, the annotations cover 30 environment labels and 47 semantic action labels for all 170 episodes, including source-supported combat labels for the 20 Gunfight cases. Seven requested visual styles cover all 200 views; the 40 Multi4 views are labeled realistic from explicit conditioning prompts. These labels describe requested styles rather than measured output appearance. The selected cameras comprise 173 third-person views (86.5\%) and 27 first-person views (13.5\%). Measured over each full recorded clip, the median camera path length is 6.11\,m, and the median maximum angular deviation from the initial orientation is $6.60^{\circ}$. Figure~\ref{fig:diversity} summarizes the interaction, style, scene-domain, camera-motion, and perspective distributions.

\begin{table}[tbp]
    \centering
    \caption{
    \textbf{Benchmark protocols and their reference information.}
    Behavior specification summarizes motion and interaction conditions;
    initial images and video context are omitted for compactness.
    State/trajectory references and task/event references are distinguished
    from conditions supplied to a model.
    \textbf{Replay} requires shared geometry and saved dynamic states sufficient
    to re-render an episode without rerunning its behavior controllers.
    \textbf{Views} denotes synchronized videos of the same world evolution;
    paired first frames or sequential perspective switches do not qualify.
    \textbf{Proxies} denotes alternative visual representations derived from
    the same states and cameras.
    \protect\ding{51}: documented support; --: not documented in the compared
    protocol or not applicable; Sim.: simulator-based evaluation, without
    confirmed recorded-state re-rendering; Subset: selected episodes.
    PROWBench combines multi-entity state and event records with replay and
    paired observations to evaluate visual realization under fixed world evolution.
    }
    \label{tab:dataset_comparison}

    \begingroup
    \providecommand{\cmark}{\ding{51}}
    \setlength{\tabcolsep}{3pt}
    \renewcommand{\arraystretch}{1.22}
    \footnotesize

    \begin{tabularx}{\linewidth}{@{}
        >{\hsize=0.95\hsize\raggedright\arraybackslash}X
        >{\hsize=1.05\hsize\raggedright\arraybackslash}X
        >{\hsize=0.95\hsize\raggedright\arraybackslash}X
        >{\hsize=1.05\hsize\raggedright\arraybackslash}X
        ccc@{}}
        \toprule
        \textbf{Benchmark}
        & \textbf{Behavior specification}
        & \textbf{State / trajectory reference}
        & \textbf{Task / event reference}
        & \textbf{Replay}
        & \textbf{Views}
        & \textbf{Proxies} \\
        \midrule

        VBench~\citep{huang2024vbench}
        & Text prompts
        & --
        & Prompt semantics
        & -- & -- & -- \\

        WorldScore~\citep{duan2025worldscore}
        & Text; camera paths
        & Target camera paths
        & Text-specified dynamics
        & -- & -- & -- \\

        WorldModelBench~\citep{li2025worldmodelbench}
        & Text instructions
        & --
        & Instructions; physics criteria
        & -- & -- & -- \\

        MIND~\citep{ye2026mind}
        & Agent navigation; camera rotation
        & Actor and camera poses
        & --
        & -- & -- & -- \\

        WorldArena~\citep{shang2026worldarena}
        & Text; robot actions
        & Robot states and actions
        & Task goals; simulator outcomes
        & Sim. & -- & -- \\

        Omni-WorldBench~\citep{wu2026omniworldbench}
        & Text interactions; optional camera paths
        & Camera targets where provided
        & Specified object effects and event order
        & -- & -- & -- \\

        MemoBench~\citep{memobench}
        & Text; optional camera paths
        & Camera poses; synthetic depth
        & Reference-video object states
        & -- & -- & -- \\

        WorldMark~\citep{xu2026worldmark}
        & Navigation via native adapters
        & Specified navigation controls
        & --
        & -- & -- & -- \\

        WBench~\citep{ying2026wbench}
        & Text interactions; navigation controls
        & Synthetic navigation paths
        & Per-turn action and event checks
        & -- & -- & -- \\

        PlayWorld~\citep{ding2026playworld}
        & Text goals; adaptive navigation
        & --
        & Goal-specific VQA rubrics
        & -- & -- & -- \\

        TriWorldBench~\citep{liu2026triworldbench}
        & Text instructions; robot actions
        & Reference robot trajectories
        & Derived action phases; task checks
        & -- & \cmark & -- \\

        \midrule
        \rowcolor{black!5}
        \textbf{PROWBench}
        & \textbf{Recorded evolution via proxies and text}
        & \textbf{Multi-entity states; camera poses}
        & \textbf{Timestamped engine events; timeline}
        & \cmark & Subset & \cmark \\

        \bottomrule
    \end{tabularx}

    \vspace{3pt}
    \parbox{\linewidth}{\scriptsize
    \textit{Scope notes.} MIND records navigation inputs and actor/camera poses,
    without structured interaction-event records in its published schema.
    Omni-WorldBench, WBench, and PlayWorld do evaluate interactions or events;
    the distinction here is the provenance of their references, not the presence
    of interaction evaluation. TriWorldBench derives coarse action phases from
    reference robot trajectories; these are not verified task-success labels.
    Engine event records specify the authored execution, not independently
    verified physical causality.}

    \endgroup
\end{table}

\subsection{\texorpdfstring{\tocheck{First-Frame Construction}}{First-Frame Construction}}
\label{app:firstframe}

\tocheck{The Verified-FF setting (Section~\ref{sec:main_results}) models the case in which an agent already has access to a concept image or character design. To provide all first-frame-conditioned methods with a common appearance reference while remaining consistent with the prescribed world state, we construct one RGB first frame for each selected view from its colored-OBB proxy and scene description. Because this image becomes part of the benchmark input, it must satisfy both appearance and geometric constraints, including the recorded person count, spatial layout, subject positions, and facing directions. We evaluated several image-generation models and selected MiniMax-H3 because it showed the strongest alignment with the proxy geometry, particularly in person count and layout. The final Verified-FF set contains 40 first frames (17 realistic and 23 stylized), all of which pass the automatic verification described below. Every method that accepts a first frame receives the same RGB image.}

\tocheck{\noindent\textbf{Generation.} Before generation, the appearance of every entity is fixed in text: each person, vehicle, and animal is assigned a concrete appearance from a reference library matched to the scene and target style, preventing the generator from implicitly choosing identities or appearances. The generation prompt combines three components: (i) a frame-0 layout contract specifying the screen position, extent, and facing direction of every visible entity according to the engine record; (ii) the proxy-control block used by the text prompts (Appendix~\ref{app:prompts}), which binds geometry to the proxy and appearance to text; and (iii) the scene appearance and style description. MiniMax-H3 receives the first 22 frames of the colored-OBB proxy together with this prompt, without an RGB image reference, and generates a short video. We use frame~0 of the generated clip, at $1280{\times}720$, as the candidate first frame. For each view-specific instance, we generate three candidates using different random seeds.}

\tocheck{\noindent\textbf{Automatic Verification.} Each candidate is subjected to hard checks against the frame-0 engine record:
\begin{itemize}
    \item \emph{Person count}: the number of persons detected by SAM~3~\citep{carion2026sam}, using the same detector as in evaluation, must equal the number of visible persons recorded by the engine.
    \item \emph{Box alignment}: the mean IoU between detected person boxes and their corresponding engine boxes must be at least $0.30$; each visible animal must individually achieve an IoU of at least $0.30$.
    \item \emph{Image defects}: Qwen3.6-27B~\citep{qwen3.6-27b} is given the candidate image together with the scene's style, world, and character descriptions, and must verify that the target style is respected and that the image contains none of the following artifacts: copied proxy geometry (e.g., wireframes, flat-colored mannequins, or boxes), large unlit black surfaces inherited from the proxy, structural defects such as detached doors, or placeholder geometry such as disc-shaped tree crowns or box-shaped props.
    \item \emph{Heading}: for vehicles and mounts with an unambiguous front direction, the visible heading (toward, away from, left of, or right of the camera) must agree with the orientation of the corresponding engine box.
\end{itemize}}

\tocheck{Candidates that fail any check are regenerated with a new seed and additional constraint sentences targeted to the failed criteria, e.g., explicitly requiring every proxy primitive to be replaced by a fully rendered object. Across regeneration rounds, we retain the best candidate that satisfies all hard checks.}

% \TODO{Describe how the text prompts are generated: source fields (scene description, entities, action timeline), template/LLM used, per-representation binding sentences, and whether prompts differ between Verified-FF and Unverified-FF and across methods.}

\subsection{\texorpdfstring{\tocheck{Text Prompts}}{Text Prompts}}
\label{app:prompts}

\tocheck{Each method receives a text prompt for every rendered view of a scene. The underlying engine provides structured metadata for each view, including scene attributes, entity annotations, temporal events, and view-dependent observations. We first construct a unified base prompt from this metadata. Except for the appearance description, all prompt content is generated deterministically from the engine records; the appearance block is the only component produced by a vision--language model. The base prompt is then converted, again deterministically, into the input format required by each method. As a result, prompt construction is reproducible for a fixed set of engine records, and differences across methods arise from their input interfaces and proxy representations. We describe below the source fields, base-prompt construction, the system prompt, representation-specific wording, and the differences between the two main-track settings and across methods. Prompts~C.1 and~C.2 at the end of this section show the base prompt and a converted prompt for one view.}

\tocheck{\noindent\textbf{Source Fields.} For each rendered view, the base prompt is constructed from the following records:
\begin{itemize}
    \item \emph{Scene description}: an authored textual description consisting of a target rendering-style line, an environment paragraph, and a protagonist paragraph. The style line is copied verbatim into every prompt. When no first frame is provided, the environment and protagonist descriptions also serve as the source of appearance information.
    \item \emph{Entity annotations}: all dynamic entities and interaction objects, together with their engine identifiers, semantic classes, and role descriptions (e.g., ``the receiving companion''). For each entity, we record whether it is visible at frame~0, based on the engine visibility flags and its projected bounding box. If an entity enters the view later, its first entry time is also recorded. In the colored-OBB representation, each entity is additionally assigned a unique identity color.
    \item \emph{Action timeline}: phase boundaries are recomputed from the recorded frames, while retaining the engine-authored sentence associated with each phase. We augment these descriptions with measured facts derived from the recorded trajectories, including the path length of each moving entity (e.g., ``moves about 10 metres along its route''), inter-entity relations, final resting locations of moving objects, and occlusion events. These descriptions do not use camera-relative direction terms.
    \item \emph{Appearance source}: in Verified-FF, appearance is derived from the verified first frame; in Unverified-FF, it is taken from the authored scene description.
\end{itemize}
Camera motion is deliberately excluded from the text prompt and is provided only through the method's control input, i.e., the proxy video or camera trajectory.}

\tocheck{\noindent\textbf{Base-Prompt Construction.} Except for the appearance block, the base prompt is generated entirely by a deterministic script, so identical engine records yield identical text. The prompt contains five components. (1) A fixed \emph{control block} specifies the roles of the two references: the proxy video provides the exact position, extent, orientation, and timing of every subject, but never its target appearance; the RGB first frame, when available, determines subject identity, appearance, and rendering style, but does not prescribe subsequent motion. The control block also defines the precedence of information sources in case of disagreement: geometry, camera, and motion follow the proxy; style and appearance follow the first frame; articulated actions follow the timestamped text. (2) \emph{Subject definitions} assign each subject a unique name and explicitly bind it to its proxy representation. (3) A one-sentence \emph{task summary} states the generation objective and shot duration. (4) A \emph{retention analysis} specifies, for each subject, which properties must be preserved from each reference. (5) A \emph{detailed description} specifies the rendering style, environment, appearance of every subject, and a contiguous action timeline covering the full five-second clip. The timeline is expressed as consecutive intervals from 00:00.000 to 00:05.000, for example: ``From 00:00.000 to 00:04.667, the protagonist walks across the road while the approaching vehicles brake and yield; the protagonist moves about 10 metres along its route; \ldots''.}

\tocheck{The appearance block is produced by Qwen3.6-27B~\citep{qwen3.6-27b} with greedy decoding. Its input consists of two images: the RGB first frame and frame~0 of the CWM proxy render. The latter is used solely as a layout reference for matching people and objects in the first frame to engine entities; it is not treated as an appearance reference. The model outputs a concise description of the opening frame and environment, the appearance and initial pose of each entity, and any discrepancy between the first frame and the recorded layout. To prevent the proxy representation from being mistaken for target appearance, proxy-specific terms such as mannequin, wireframe, and bounding box are disallowed.}

\tocheck{Every composed prompt is checked by an automatic validator. The validator requires that (i) every subject is defined, preserved, and described; (ii) only the allowed references are mentioned; (iii) the timeline is contiguous and covers exactly 0--5\,s; (iv) prompts without a first frame contain no first-frame references; and (v) the required representation-specific wording is present. If validation fails, the validation errors are returned to the VLM for one regeneration attempt. }

\tocheck{\noindent\textbf{System Prompt.} The control block plays the role of a system prompt, but how it reaches each model depends on the model's text interface. CWM is the only method whose interface has a separate system role. Its text encoder applies the Qwen chat template with two turns, a fixed system prompt followed by a user turn containing the \textless Picture~1\textgreater{} and \textless Video~1\textgreater{} references and the caption, and encodes the formatted sequence in a single pass. 
The release ships two such system prompts, one for the first generation window and one for continuation windows. Our control block is adapted from this system prompt, with three changes for the box proxy: the proxy gives each subject's exact extent and must be followed precisely (CWM's prompt treats its proxy as a coarse reference), the white face of a box marks the entity's front, and the hexadecimal colour tags are identity codes. MiniMax-H3 and LynnReal-Omni use no chat template and no system role: their text encoder receives one token sequence consisting of a label and the visual tokens of each reference (\textless Picture~1\textgreater{} for the first frame, \textless Video~1\textgreater{} for the proxy video), followed by the prompt verbatim, so the control block enters as the first part of the prompt (Prompt~C.1). The Seedance~2.5 API likewise accepts a single prompt. In Unverified-FF, Seedance~2.5 receives the base prompt with the control block at its start, as MiniMax-H3 does. In Verified-FF, it receives a deterministic conversion of the same base prompt into its reference syntax, as C2R and CWM receive conversions into theirs: the conversion keeps the style line, the appearance of every subject, the binding of each subject to its box by colour name and legend colour, the proxy semantics, and the timestamped timeline, but states the proxy semantics as ordinary sentences instead of a control block (Table~\ref{tab:prompt_methods}). C2R, Cosmos-Transfer2.5, and the camera-conditioned world models receive no system prompt, and their converted prompts contain no proxy terminology (Prompt~C.2).}

\tocheck{\noindent\textbf{Representation-Specific Wording.} Only two parts of the base prompt depend on the proxy representation: the sentence defining the semantics of the proxy video, and the rule used to bind each subject to its proxy.
\begin{itemize}
    \item \emph{Coarse 3D proxy:} the full scene is represented as an untextured light-grey low-poly proxy. People are articulated grey block figures whose recorded limb motion must be preserved. All grey surfaces encode geometry only and must not be interpreted as target colors or materials.
    \item \emph{CWM proxy}: people are represented as flat blue upright mannequins with a fixed pose; they translate and rotate but carry no limb-motion information. Interaction objects are shown as red boxes. The environment is rendered using flat semantic colors that act only as proxy codes and do not correspond to target materials or colors.
    \item \emph{Colored-OBB:} people, vehicles, animals, and interaction objects are represented as semi-transparent solid boxes, with one unique identity color per entity. The color serves only to identify which box corresponds to which subject and must not be interpreted as a target color or material. The white face marks the front of the entity; hence, if the white face is not visible, the entity is facing away from the camera. The white face itself is likewise only a proxy marker and must not appear in the generated result.
\end{itemize}}

\tocheck{For colored-OBBs, each subject definition explicitly identifies its proxy using the hexadecimal legend color. For example: ``\textless Subject~3\textgreater{} is the passenger car: in \textless Picture~1\textgreater{} a classic red sedan with a white license plate, parked on the right side of the street facing away from the camera; in \textless Video~1\textgreater{} the object proxy tagged \#f79e36, whose white face is its front.'' Subjects that are not visible at frame~0 are additionally annotated with their first entry time. In the main track, the methods that receive the box proxy (MiniMax-H3, LynnReal-Omni and Seedance~2.5) use the colored-OBB prompt; C2R, which receives the coarse 3D proxy, and CWM, which receives the CWM proxy, use converted prompts without proxy wording (Table~\ref{tab:prompt_methods}). The coarse 3D and CWM proxy base prompts are used for MiniMax-H3 only in the representation study (Table~\ref{tab:h3_representations}).}

\tocheck{\noindent\textbf{Verified-FF versus Unverified-FF.} Table~\ref{tab:prompt_settings} summarizes the two main-track settings. They share the same structural skeleton, including subject definitions, proxy bindings, the action timeline, and all measured facts. They differ only in the source of appearance information and in the role of the first frame.}

\begin{table}[h]
\centering
\caption{\tocheck{\textbf{Base prompt in the two main-track settings.}}}
\label{tab:prompt_settings}
\tablestyle{3pt}{1.15}\scriptsize
\begin{tabularx}{\linewidth}{@{}l>{\raggedright\arraybackslash}X>{\raggedright\arraybackslash}X@{}}
\toprule
 & \textbf{Verified-FF} & \textbf{Unverified-FF} \\
\midrule
References & \textless Picture~1\textgreater{} (first frame) and \textless Video~1\textgreater{} (proxy) & \textless Video~1\textgreater{} only \\
Task & keyframe completion and reference generation; the generated clip must start exactly from the provided first frame & reference generation \\
Appearance source & Qwen3.6-27B description of the verified first frame & authored scene description \\
Style source & authored style line; the first frame also visually exhibits this style & authored style line \\
Control rule & for overall style and the appearance of subjects visible in the first frame, the first frame takes precedence over text & the appearance of every subject is specified by text \\
Structural skeleton & \multicolumn{2}{l}{identical: subject definitions, proxy bindings, action timeline, and measured facts} \\
\bottomrule
\end{tabularx}
\end{table}

\tocheck{\noindent\textbf{Prompt Formats across Methods.} The base prompt is designed for models that can consume explicitly tagged image and video references. For methods with different interfaces, we apply deterministic transformations that modify the format while preserving the underlying semantic content to the extent permitted by the target interface (Table~\ref{tab:prompt_methods}). No transformation relies on a language model.  In Unverified-FF, the appearance description used by CWM is derived from frame~0 of the MiniMax-H3 output, which also serves as CWM's RGB anchor frame.}

\begin{table}[h]
\centering
\caption{\tocheck{\textbf{Prompt format by method.}}}
\label{tab:prompt_methods}
\tablestyle{3pt}{1.25}\scriptsize\renewcommand{\baselinestretch}{1.25}\selectfont
\begin{tabularx}{\linewidth}{@{}>{\raggedright\arraybackslash}p{2.6cm}>{\raggedright\arraybackslash}X@{}}
\toprule
\textbf{Methods} & \textbf{Prompt format} \\
\midrule
MiniMax-H3, LynnReal-Omni & base prompt verbatim, including the control block, tagged references, subject bindings with legend colors, and the complete timestamped timeline \\
\addlinespace[3pt]
Seedance~2.5 & Verified-FF: converted to Seedance's reference syntax (\texttt{@Image1}, \texttt{@Video1}); retains the style line, the appearance of every subject, a ``who is who'' list identifying each box by color name and legend color, proxy semantics, the timestamped timeline, and field-of-view information; removes the control block, subject tags, retention analysis, and soundscape. Unverified-FF: base prompt with the control block, as for MiniMax-H3 (see \emph{System prompt}) \\
\addlinespace[3pt]
C2R, Cosmos-Transfer2.5, EchoWM, SANA-WM, LingBot~2.0, AlayaWorld~v1.1 & converted to the single-paragraph caption style used to train C2R: an opening description of the style, setting, number of people, and appearance of every subject, followed by the action progression in prose without timestamps and a final quality sentence; removes tags, proxy terminology, and camera-specific text, and is shortened to satisfy C2R's 512-token text limit \\
\addlinespace[3pt]
CWM & converted to CWM's caption format, consisting of \emph{appearance and scene}, a timestamped \emph{protagonist action timeline}, and \emph{other visible actors}; removes subject tags and proxy-specific wording, while CWM adds its own system prompt for the mixed proxy render and RGB anchor frame \\
\bottomrule
\end{tabularx}
\end{table}

\begin{tcolorbox}[breakable,enhanced,colback=black!2,colframe=black!35,boxrule=0.4pt,arc=1pt,left=4pt,right=4pt,top=3pt,bottom=3pt,fontupper=\scriptsize\rmfamily,title={\scriptsize\sffamily\bfseries Prompt C.1\quad Base prompt, Verified-FF, \textcolor{gray}{[\ldots]} marks omitted text},coltitle=black,colbacktitle=black!8]
\textbf{Control block (placed at the start of the prompt; plays the role of a system prompt)}\par
AWM\_PROXY\_CONTROL. Two visual references accompany this text. \textless{}Video 1\textgreater{} is the conditioning proxy video: treat it as the exact spatiotemporal reference for where every subject is, how large it is and how it moves, but never as an appearance template or literal target geometry. \textless{}Picture 1\textgreater{} is the global RGB anchor image: it defines identity and appearance, never motion. \textless{}Picture 1\textgreater{} is the ABSOLUTE standard for the worldview, art direction and overall style of the entire take, and on worldview, art direction and style it outranks even the description below. It provides NOTHING about any later frame's camera, composition, motion or interactions. Each subject's position and motion come from \textless{}Video 1\textgreater{}. A subject's appearance follows \textless{}Picture 1\textgreater{} when that subject is visible in it; subjects not visible in \textless{}Picture 1\textgreater{} take their appearance from the text, rendered consistently in \textless{}Picture 1\textgreater{}'s worldview and art direction.
\par
Human proxies in \textless{}Video 1\textgreater{} provide only framewise root position, occupied scale (in this representation the box is the entity's exact extent), and whole-body root orientation, which are exact and must be followed precisely. \textcolor{gray}{[\ldots]} On every human, vehicle and animal proxy in \textless{}Video 1\textgreater{} the white face is the front of that entity's body: white there is a direction code, never an object, a material or a light, and it must not be painted. The hex colour tags used below only say which proxy stands for which subject: they are identity codes, never colours or materials of the result.
\par
\textcolor{gray}{[\ldots]} \textit{(further proxy rules; vehicle proxies; environment cues)}
\par
When sources conflict, resolve in this order: (1) geometry, camera, position and motion follow \textless{}Video 1\textgreater{}; (2) the worldview, art direction and overall style follow \textless{}Picture 1\textgreater{} absolutely - above the description below; identity, appearance and materials follow \textless{}Picture 1\textgreater{} for subjects visible in it, and follow the text for subjects not visible in it; (3) the timestamped text defines the concrete events, actions and rewrite targets within those constraints.
\par
Clip contract: this clip is the very beginning of the take - nothing precedes it. \textless{}Picture 1\textgreater{} is the global appearance anchor for the whole take, and the first frame of the target is locked to this exact image: the clip begins at precisely this visual state - identities, appearance, materials, art direction, lighting, camera framing and layout - and continues directly from it while performing the timestamped actions. Do not restart, redesign or re-interpret any identity.
\par\medskip
\textbf{subject\_definitions:}\par
\textless{}Picture 1\textgreater{} is the first frame of [Shot 1]: the opening frame of the target video, already rendered in the target art style of this take, showing A man in a dark suit walks down the center of a city street, flanked by two parked cars and three pedestrians standing on the sidewalks, exactly as they must appear at 00:00.000.
\par
\textless{}Video 1\textgreater{} is the conditioning proxy video of the same 5-second shot (its role is fixed by the control block above). The clip contains exactly 4 human proxies standing for \textless{}Subject 1\textgreater{} (the protagonist), \textless{}Subject 5\textgreater{} (independently moving adult participant 1), \textless{}Subject 6\textgreater{} (independently moving adult participant 2), \textless{}Subject 7\textgreater{} (independently moving adult participant 3) and 2 object proxies standing for \textless{}Subject 3\textgreater{} (the passenger car), \textless{}Subject 4\textgreater{} (the passenger car); nothing else in \textless{}Video 1\textgreater{} is a person or an interaction object.
\par
\textless{}Subject 1\textgreater{} is the protagonist: in \textless{}Picture 1\textgreater{} A man with short brown hair, wearing a dark charcoal suit and black shoes, walking away from the camera down the center of the road; in \textless{}Video 1\textgreater{} the human proxy tagged \#366ef7, whose white face is its front.
\par
\textless{}Subject 3\textgreater{} is the passenger car: in \textless{}Picture 1\textgreater{} A classic red sedan with a white license plate, parked on the right side of the street facing away from the camera; in \textless{}Video 1\textgreater{} the object proxy tagged \#f79e36, whose white face is its front.
\par
\textless{}Subject 4\textgreater{} is the passenger car: in \textless{}Picture 1\textgreater{} A classic black sedan with chrome bumpers and round headlights, parked on the left side of the street facing the camera; in \textless{}Video 1\textgreater{} the object proxy tagged \#6636f7, whose white face is its front.
\par
\textless{}Subject 5\textgreater{} is independently moving adult participant 1: in \textless{}Picture 1\textgreater{} A young man with dark hair, wearing a mustard hooded jacket, dark shorts, and white sneakers, standing on the left sidewalk facing the street; in \textless{}Video 1\textgreater{} the human proxy tagged \#a7f736 that starts on the left of the frame, whose white face is its front.
\par
\textless{}Subject 6\textgreater{} is independently moving adult participant 2: in \textless{}Picture 1\textgreater{} A person with short blonde hair, wearing a maroon hoodie and blue jeans, standing on the right sidewalk facing the street; in \textless{}Video 1\textgreater{} the human proxy tagged \#f736df that starts on the right of the frame, whose white face is its front.
\par
\textless{}Subject 7\textgreater{} is independently moving adult participant 3: in \textless{}Picture 1\textgreater{} A man with dark hair, wearing a white t-shirt under dark blue overalls and brown shoes, standing on the left sidewalk near the black car; in \textless{}Video 1\textgreater{} the human proxy tagged \#36f7d6 that starts on the left of the frame, whose white face is its front.
\par
\textcolor{gray}{[\ldots]} \textit{(Subject 2: environment)}
\par\medskip
\textbf{summary:}\par
[keyframe completion + reference generation] Starting exactly from \textless{}Picture 1\textgreater{} as the first frame of [Shot 1], generate a 5.00-second continuous shot in \textless{}Subject 2\textgreater{} of \textless{}Subject 1\textgreater{} crossing a road as approaching vehicles brake and yield, involving \textless{}Subject 3\textgreater{} and \textless{}Subject 4\textgreater{} and \textless{}Subject 5\textgreater{} and \textless{}Subject 6\textgreater{} and \textless{}Subject 7\textgreater{}. Use \textless{}Video 1\textgreater{} for layout, root motion, object trajectories, action timing and the camera path, and keep the appearance, materials and lighting established in \textless{}Picture 1\textgreater{}.
\par\medskip
\textbf{retention\_analysis:}\par
\textless{}Picture 1\textgreater{} ([Shot 1] first frame): fully\_preserved - the first frame of the target video reproduces this image exactly, including every person's appearance, the objects, the scene layout, materials, lighting and framing; later frames evolve continuously from it.
\par
\textless{}Subject 1\textgreater{} (appears in [Shot 1]): partially\_preserved - retain the proxy's framewise root position, occupied scale, root orientation (the white face of the box is the front of the body), identity continuity and interaction timing from \textless{}Video 1\textgreater{}; reconstruct its appearance from \textless{}Picture 1\textgreater{} and its articulated motion from the timeline below.
\par
\textcolor{gray}{[\ldots]}
\par\medskip
\textbf{detailed\_description:}\par
Anime-style rendering: a 2D anime look with hand-drawn character design and painted backgrounds over the full 3D world --- modern digital anime: gradient shading, glossy highlights, clean thin contour lines, vivid high-contrast colours; \textcolor{gray}{[\ldots]}
\par
[Shot 1] The shot opens exactly on \textless{}Picture 1\textgreater{} and evolves continuously from that frame. \textcolor{gray}{[\ldots]} \textit{(setting and appearance of every subject, as in subject\_definitions)} From 00:00.000 to 00:04.667, the protagonist walks across the road while the approaching vehicles brake and yield; the protagonist moves about 10 metres along its route; independently moving adult participant 1 moves about 5 metres along its route; independently moving adult participant 2 moves about 4 metres along its route; independently moving adult participant 3 moves about 4 metres along its route; \textless{}Subject 3\textgreater{} moves about 9 metres along its route; \textless{}Subject 4\textgreater{} moves about 6 metres along its route. From 00:04.667 to 00:05.000, the protagonist reaches the far side of the road and keeps walking; independently moving adult participant 1 moves a short distance along its route; independently moving adult participant 2 moves a short distance along its route; independently moving adult participant 3 moves a short distance along its route; \textless{}Subject 3\textgreater{} moves about half a metre along its route; \textless{}Subject 4\textgreater{} moves a short distance along its route. \textless{}Subject 3\textgreater{} is one continuous object throughout and ends the clip about 7 metres from the protagonist. \textless{}Subject 4\textgreater{} is one continuous object throughout and comes to rest about 5 metres from the protagonist. Keep the reference's fixed 53-degree vertical field of view. \textcolor{gray}{[\ldots]}
\par\medskip
\textbf{overall\_soundscape:}\par
The ambient hum of a quiet city street with distant traffic noise and the rhythmic sound of footsteps on asphalt.
\par\medskip
\textbf{non\_diegetic\_music:}\par
N/A
\end{tcolorbox}

\begin{tcolorbox}[breakable,enhanced,colback=black!2,colframe=black!35,boxrule=0.4pt,arc=1pt,left=4pt,right=4pt,top=3pt,bottom=3pt,fontupper=\scriptsize\rmfamily,title={\scriptsize\sffamily\bfseries Prompt C.2\quad Converted prompt for C2R, Cosmos-Transfer2.5 and the camera-conditioned world models},coltitle=black,colbacktitle=black!8]
A high-quality video of a person crossing a road as approaching vehicles brake and yield. Anime-style rendering: a 2D anime look with hand-drawn character design and painted backgrounds over the full 3D world --- modern digital anime: gradient shading, glossy highlights, clean thin contour lines, vivid high-contrast colours; no proxy colours, wireframes, block figures or game UI; the opening frame establishes this look and every later frame keeps it. The setting is a wide urban street paved with asphalt and marked with white lane lines, flanked by sidewalks with moss-toned metal railings. There are 4 persons in this video. The protagonist is a man with short brown hair, wearing a dark charcoal suit and black shoes, walking away from the camera down the center of the road. The passenger car 1 is a classic red sedan with a white license plate, parked on the right side of the street facing away from the camera. The passenger car 2 is a classic black sedan with chrome bumpers and round headlights, parked on the left side of the street facing the camera. Independently moving adult participant 1 is a young man with dark hair, wearing a mustard hooded jacket, dark shorts, and white sneakers, standing on the left sidewalk facing the street. Independently moving adult participant 2 is a person with short blonde hair, wearing a maroon hoodie and blue jeans, standing on the right sidewalk facing the street. Independently moving adult participant 3 is a man with dark hair, wearing a white t-shirt under dark blue overalls and brown shoes, standing on the left sidewalk near the black car. Over the five seconds, the protagonist walks across the road while the approaching vehicles brake and yield, and then the protagonist reaches the far side of the road and keeps walking. Human motion is natural and continuous, presented in a cinematic style with sharp details, clean textures, and no blur or visual artifacts.
\end{tcolorbox}

\clearpage
\section{Metric Details}
\label{appendix_metric}

\paragraph{Entity control.}
We evaluate entities marked as dynamic or interaction candidates, excluding terrain, effects, and scene geometry. Entities that are not visible in the reference view are excluded from evaluation, with additional filtering for first-person views to remove the camera wearer and other entities that dominate the field of view.
Generated entities are localized using SAM~3~\citep{carion2026sam} from text prompts derived from their semantic labels. 
%
% If SAM~3 fails to detect a label throughout a clip, we fall back to SAM~2~\citep{sam2}, initialized from the reference box.
%
Generated tracks are matched to reference entities by label-compatible Hungarian matching using track-level mean IoU.

For entity $j$ with visible instants $\mathcal{V}_j$,
\begin{equation}
\mathrm{IoU}_j = \frac{1}{|\mathcal{V}_j|}\sum_{t\in\mathcal{V}_j}\mathrm{IoU}\big(\hat b_j(t), b_j(t)\big),\qquad
\mathrm{CErr}_j = \frac{1}{|\mathcal{M}_j|}\sum_{t\in\mathcal{M}_j}\frac{\lVert \hat c_j(t)-c_j(t)\rVert_2}{\sqrt 2},
\label{eq}
\end{equation}
where $b_j(t)$ and $\hat b_j(t)$ denote the reference and generated bounding boxes in normalized image coordinates, and $\mathcal{M}_j\subseteq\mathcal{V}_j$ contains the instants at which the generated track is detected. Missing boxes contribute zero IoU, so \textbf{BBox IoU} captures both localization accuracy and persistence, while \textbf{Center Error} measures localization error conditional on detection and is normalized by the image diagonal.
Clip-level BBox IoU is averaged over all evaluated entities, whereas Center Error is averaged over entities with at least one matched instant. Clips with no evaluable entities are excluded.

\paragraph{Camera control.}
Camera trajectories of generated videos are recovered with ViPE~\citep{huang2025vipe} in pose-only mode with intrinsics fixed to the recorded camera (after the method's crop and resize) and with dynamic entities masked using their engine labels. Poses are converted to the engine convention and expressed relative to the first instant. Following CamI2V~\citep{zheng2024cami2v}, with $R_t$, $T_t$ the reference and $\hat R_t$, $\hat T_t$ the estimated relative poses on the grid $\mathcal{G}$,
\begin{equation}
\begin{aligned}
    \mathrm{RotErr} &= \frac{1}{|\mathcal{G}|}\sum_{t\in\mathcal{G}} \arccos\frac{\operatorname{tr}(\hat R_t R_t^{\top})-1}{2}, \qquad
    \mathrm{TransErr} = \frac{1}{|\mathcal{G}|}\sum_{t\in\mathcal{G}} \big\lVert \bar T_t - \bar{\hat T}_t \big\rVert_2,\\
    \mathrm{CamMC} &= \frac{1}{|\mathcal{G}|}\sum_{t\in\mathcal{G}} \big\lVert [R_t\,|\,\bar T_t] - [\hat R_t\,|\,\bar{\hat T}_t] \big\rVert_F ,
\end{aligned}
\label{eq:camera_metrics}
\end{equation}
where each translation sequence is divided by its own maximum norm, $\bar T_t = T_t/(\max_s\lVert T_s\rVert+10^{-9})$, to remove the scale ambiguity of monocular estimation. 
%
%RotErr is reported in degrees. 
RotErr is reported in degrees in all tables.
The normalization also removes differences in the overall magnitude of camera motion, so TransErr and CamMC measure the shape of the path rather than its length. 
% Because the estimator is itself imperfect, the engine renders are scored in the same way; their errors indicate the resolution of the camera columns but are not strict error bounds.

\paragraph{Visual and temporal quality.}
\textbf{Imaging Quality} (MUSIQ~\citep{ke2021musiq}) and \textbf{Aesthetic Quality} (the LAION aesthetic predictor) follow VBench~\citep{huang2024vbench}. \textbf{Warp Error}~\citep{lai2018learning,liu2024evalcrafter} estimates RAFT optical flow between consecutive frames at $672\times384$, warps the next frame back to the current one, masks occlusions by forward--backward consistency and reports the masked mean squared RGB error ($\times10^{-3}$). \textbf{Motion Smoothness} is the VBench score that reconstructs frames with a video frame-interpolation model~\citep{li2023amt}. \textbf{CLIP-Score}~\citep{radford2021clip} is the mean cosine similarity ($\times100$) between CLIP ViT-B/32 embeddings of the 40 grid frames and the episode's action title. These metrics are computed without reference to the engine state and therefore measure plausibility, not adherence.

\paragraph{Appearance Compliance and Consistency.} For each intended participant in the multi-view setting, \emph{Appearance Compliance} (\emph{Comp.}) asks whether the person at the engine-projected location matches the authored appearance description shared by all methods, judged per view at 0.5, 2.5, and 4.5\,s from the full frame and a target crop. \emph{Cross-view Appearance Consistency} (\emph{Cons.}) asks whether the participant looks the same across views: the judge compares every pair of synchronized views without the description, ignoring changes in pose, viewpoint, lighting, and front/back visibility. A participant can thus be consistent yet non-compliant. Both metrics use Qwen3.6-27B with greedy decoding on five attributes (clothing color, clothing style, hair, accessories, and build) and skip attributes that are undescribed or not visible. An observation $o$ (person--view--time for Comp., person--view-pair--time for Cons.) with judgeable attributes $\mathcal{A}_o$ scores
\begin{equation}
    q_o=\prod_{a\in\mathcal{A}_o}\mathbf{1}[\text{attribute }a\text{ matches or agrees}],
    \label{eq:appearance_pass}
\end{equation}
i.e., every judged attribute must pass. Observations with an absent target or no judgeable attribute are unscored. Scores are averaged over views or view pairs, then over participant--time groups, then equally over scenes, and reported in percent; because unscored observations are excluded, both metrics should be read together with the BBox miss rate.

%\emph{Judgeable coverage} uses the same hierarchy with a 1 for a scored observation and 0 otherwise. Unlike the event scores, unjudgeable appearance observations reduce coverage rather than count as failures. These conditional percentages should therefore be read with coverage and BBox miss rate. 
%The judgments have not been systematically calibrated against human annotations; localization drift can also affect the appearance assessment.  

% \paragraph{Geometric multi-view consistency.}
% We also report \textbf{MEt3R~\citep{asim2025met3r}}, which compares image features after aligning views through geometry predicted from the generated frames. It measures mutual consistency of the generated views, not agreement with the engine geometry, and complements the person-level appearance judgments rather than measuring adherence to an appearance description. We evaluate all six view pairs at nine synchronized times per scene, average both comparison directions, and average scenes equally; lower is better.

\paragraph{Geometric multi-view consistency.}
We also report \textbf{MEt3R~\citep{asim2025met3r}}, which compares image features after aligning views through geometry predicted from the generated frames. It measures mutual consistency of the generated views, not agreement with the engine geometry, and complements the person-level appearance judgments rather than measuring adherence to an appearance description. We evaluate all six pairs of the four views at nine synchronized times per scene, average both comparison directions, and average scenes equally; lower is better.

% \subsection{\texorpdfstring{\tocheck{First-Frame Construction}}{First-Frame Construction}}
% \label{app:firstframe}
% \tocheck{The Verified-FF setting (Section~\ref{sec:main_results})} models an agent that already has a concept image or character design. To give every method that accepts a first frame the same appearance reference, consistent with the prescribed world state, we generate one RGB first frame per pair from the colored-OBB proxy of that camera and the scene description. Because this image is part of the benchmark, it must also match the recorded person count, positions, facing and layout; after comparing several image generators, we use MiniMax-H3, whose first frames aligned best with the proxy geometry in person count and layout. The set contains 17 realistic and 23 stylized first frames, each admitted only after automatic checks of person count, box alignment, style and facing. All methods that accept a first frame receive the identical image.

\clearpage
\section{\texorpdfstring{\tocheck{Experiments}}{Experiments}}
\label{app:experiments}

\tocheck{This appendix gives implementation details of the evaluated models and qualitative examples for Section~\ref{sec:experiments}.}

\subsection{Model Configurations and Implementation Details}
\label{app:model_configs}

Table~\ref{tab:method_specs} summarizes the conditioning inputs, spatial resolutions, frame rates, and delivered frame counts of the evaluated methods. The specifications apply to the five-second main-track experiments; PWM is evaluated only on the gunfight set. Spatial dimensions are reported as width $\times$ height. Input sizes describe the supplied conditioning files, with internal resizing stated where applicable; output sizes and frame counts describe the delivered videos.
EchoWM~\citep{zhang2026echowm} uses the Base checkpoint, while SANA-WM~\citep{zhu2026sana} and LingBot-World~2.0~\citep{robbyant2026lingbotworld} use their bidirectional checkpoints; all three generate the full clip without autoregressive rollout. For AlayaWorld~v1.1~\citep{team2026alayaworldv11}, we use the camera-controlled Stage2b AR teacher with the Stage3 DMD adapter, using four denoising steps per rollout round, because the released Stage1 bidirectional checkpoint supports first-frame and text conditioning but does not provide camera-trajectory control.
% footnote{Release documentation checked on September 29, 2026: \href{https://huggingface.co/AlayaLab/AlayaWorld-stage1}{Stage1 bidirectional checkpoint} and \href{https://huggingface.co/AlayaLab/AlayaWorld-v1.1-stage3}{v1.1 DMD adapter}.}

% EchoWM uses the Base checkpoint, and SANA-WM uses the bidirectional version; both generate the clip without autoregressive rollout. The current LingBot-World~2.0 results also use the bidirectional checkpoint, whereas AlayaWorld~v1.1 uses chunk-autoregressive generation.

\begin{table}[!ht]
\centering
\caption{\textbf{Evaluated Model Configurations and Input--Output Specifications.} \tocheck{Figure~\ref{fig:test_condition} shows the full test condition of one Unverified-FF sample, and Figure~\ref{fig:track_conditions} the conditions of the other settings.}  }
\label{tab:method_specs}
\begingroup
\fontsize{8}{10}\selectfont
\setlength{\tabcolsep}{4pt}
\renewcommand{\arraystretch}{1.2}
\begin{tabularx}{\linewidth}{@{}>{\raggedright\arraybackslash}p{0.17\linewidth}>{\raggedright\arraybackslash}X>{\raggedright\arraybackslash}p{0.25\linewidth}ccc@{}}
\toprule
\textbf{Method} & \textbf{Conditioning} & \textbf{Supplied spatial input} & \textbf{Output} & \textbf{FPS} & \textbf{Frames} \\
\midrule
% \multicolumn{6}{@{}l}{\emph{Camera-conditioned world models}} \\
EchoWM~\citep{zhang2026echowm}  & RGB first frame, camera trajectory & $1280\times720$ image & $1280\times704$ & 24 & 121 \\
SANA-WM~\citep{zhu2026sana} & RGB first frame, camera trajectory& $1280\times720$ image & $1280\times704$ & 16 & 80 \\
LingBot-World~2.0~\citep{robbyant2026lingbotworld} & RGB first frame, camera trajectory & $1280\times720$ image & $1280\times720$ & 16 & 81 \\
AlayaWorld~v1.1~\citep{team2026alayaworldv11} & RGB first frame, camera trajectory; chunk-autoregressive generation, 4 rounds, 4 denoising steps per round & $1280\times720$ image, resized and cropped to $960\times544$ & $960\times544$ & 24 & 120 \\
\midrule
% \multicolumn{6}{@{}l}{\emph{Proxy-video-to-RGB models}} \\
Cosmos-Transfer2.5~\citep{ali2025world} &  Depth, class-level segmentation & $1280\times720$, 24\,fps, 121 frames & $1280\times720$ & 24 & 121 \\
C2R~\citep{gomez2026coarse} & Coarse 3D proxy video & $832\times480$, 16\,fps, 81 frames & $832\times480$ & 16 & 81 \\
CWM~\citep{chen2026code} & RGB first frame, CWM proxy, depth & First frame $1280\times720$; CWM proxy and depth $336\times192$ & $1344\times768$ & 24 & 120  \\
PWM~\citep{huang2026programmableworldmodel} & RGB first frame, instance-ID maps from 3D boxes & $832\times480$ & $832\times480$ & 16 & 81 \\
LynnReal-Omni~\citep{mao2026lynnrealomninativemultimodalvideo} & Colored-OBB proxy video; first frame as reference & $1280\times720$, encoded at $1344\times768$ & $1344\times768$ & 24 & 120 \\
\midrule
% \multicolumn{6}{@{}l}{\emph{Video foundation models}} \\
MiniMax-H3~\citep{minimaxh3} & Colored-OBB proxy video; first frame as reference & $1344\times768$, 24\,fps, 124 frames & $1344\times768$ & 24 & 120 \\
Seedance~2.5~\citep{seedance25} & Colored-OBB proxy video; first frame as reference  & $1280\times720$, 30\,fps, 150 frames & $1280\times720$ & 24 & 121 \\
\bottomrule
\end{tabularx}

\endgroup
\end{table}

\begin{figure*}[t]
    \centering
    \includegraphics[width=\textwidth]{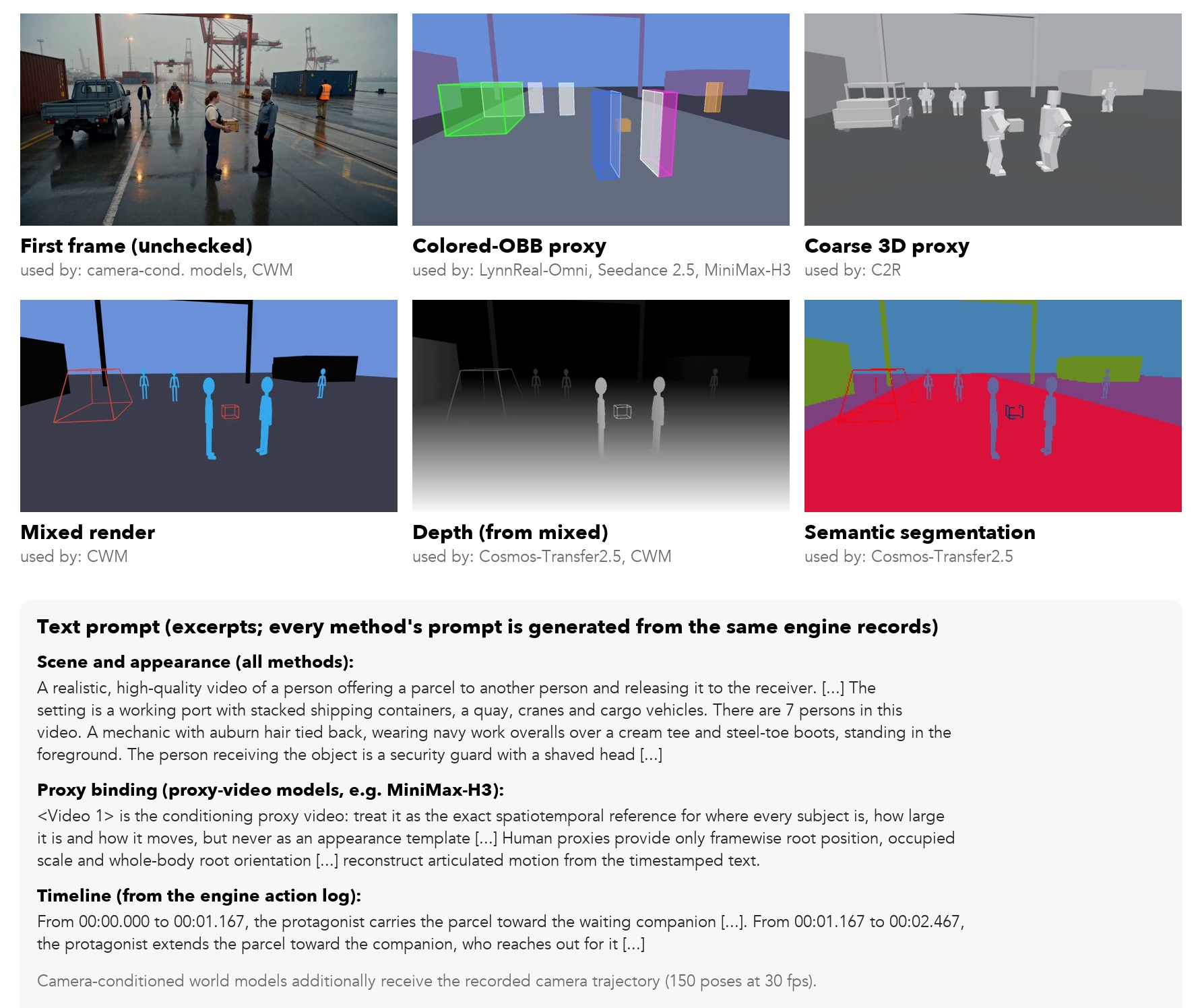}
    \caption{\textbf{\tocheck{Test condition of one sample.}} \tocheck{Unverified-FF, handoff scene (follow camera): every conditioning input provided for this sample, the methods that receive it (Table~\ref{tab:method_specs}), and excerpts of the text prompt. Proxy frames are shown at 0.5\,s; the first frame is frame~0 of the MiniMax-H3 output, which is not checked against the recorded state.}}
    \label{fig:test_condition}
\end{figure*}

\begin{figure*}[t]
    \centering
    \includegraphics[width=\textwidth]{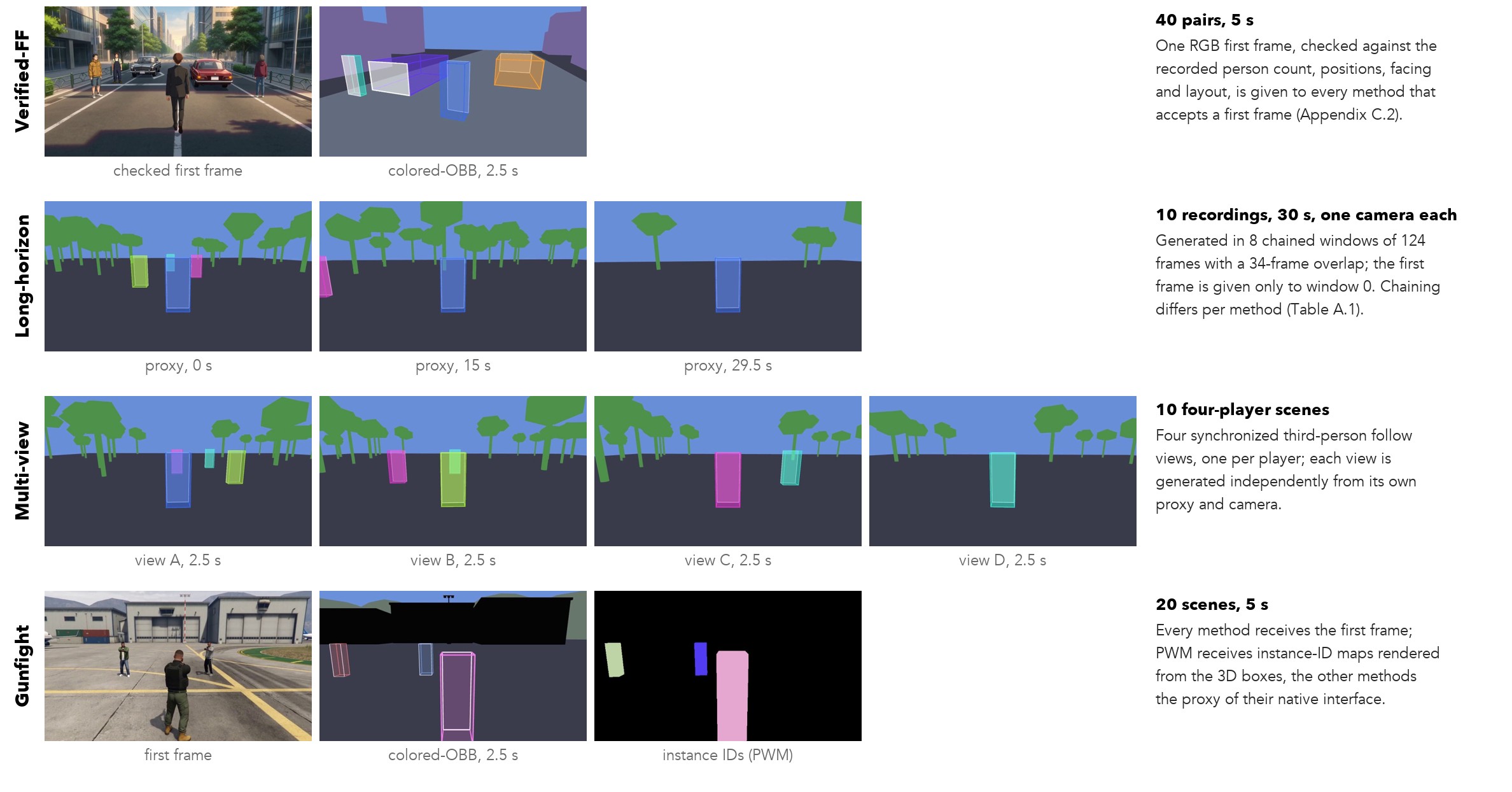}
    \caption{\textbf{\tocheck{Test conditions of the other settings.}} \tocheck{One example per setting: Verified-FF (main track), long-horizon and multi-view (challenge track), and the gunfight set used for the PWM comparison. Compared with the Unverified-FF example in Figure~\ref{fig:test_condition}, the settings differ in the first frame, the clip length and generation protocol, and the number of synchronized views. All methods also receive a text prompt generated from the engine records.}}
    \label{fig:track_conditions}
\end{figure*}

\subsection{\texorpdfstring{\tocheck{Qualitative Examples}}{Qualitative Examples}}
\label{app:qual}

\begin{figure*}[h]
    \centering
    \includegraphics[width=\textwidth]{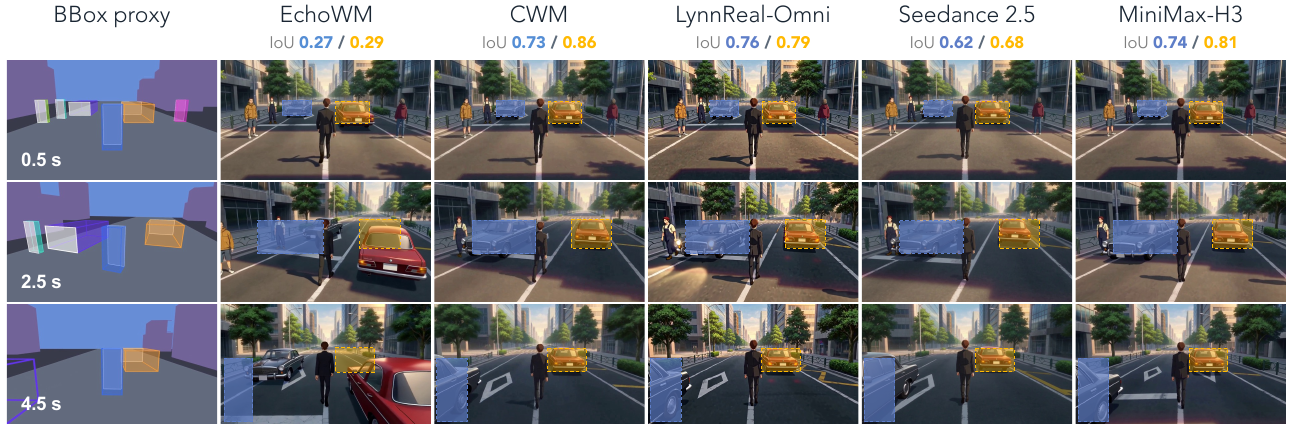}
    \caption{\textbf{Entity-level control.} A pedestrian walking down a city street,
third-person follow camera, same first frame for all methods. The semi-transparent white face on each proxy box marks the entity's front. Dashed: engine reference boxes of the oncoming car
(cyan) and the car ahead (orange). The proxy-conditioned models keep both cars in their boxes (IoU 0.62–0.86);
EchoWM, conditioned on the camera trajectory only, does not (0.27 and 0.29).
    }
    \label{fig:dv200_firstframe_qual}
\end{figure*}

\begin{figure*}[p]
    \centering
    \includegraphics[width=0.95\textwidth]{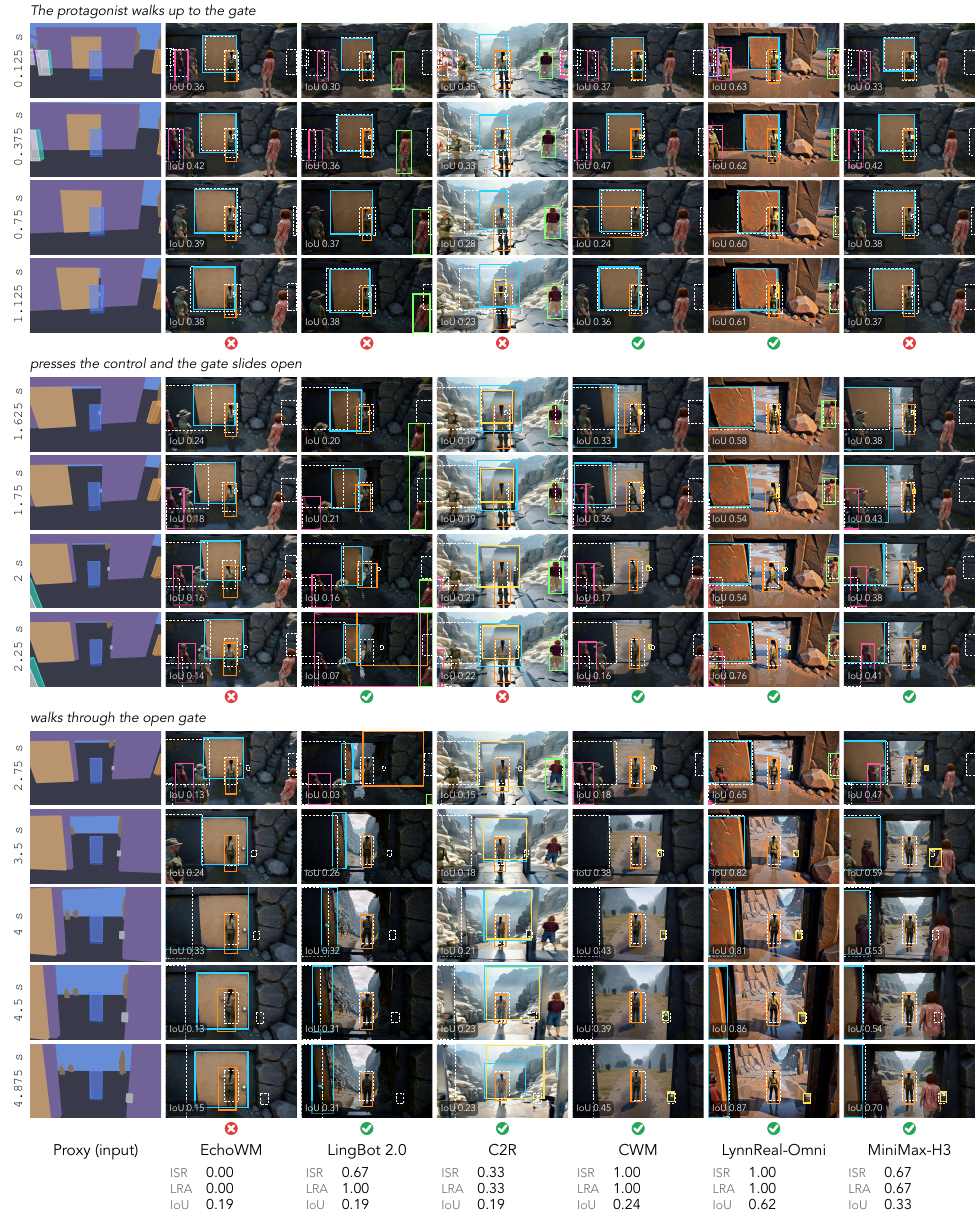}
    \caption{\textbf{\tocheck{Interaction Success Rate (ISR) and Logic-Render Alignment (LRA) on one scene.}} \tocheck{Unverified-FF, gate scene. Columns: the colored-OBB proxy and six methods; rows: the three recorded events (approach, opening, passed-through), four to five frames each. Badges below each event: whether the judge, which sees frames from the whole event window, finds the event realized, i.e., both its action and its end state are judged present. Dashed white boxes are the engine reference boxes and solid boxes the boxes detected in the generated video (one color per entity); the overlay gives the mean BBox IoU of each frame. Bottom: ISR, the fraction of realized events, LRA, the fraction of timeline segments whose action is depicted, and the BBox IoU of the whole video. The two can differ, e.g., LingBot~2.0 depicts all three actions (LRA $1.00$) but misses the end state of the first event (ISR $0.67$).}}
    \label{fig:metric_illustration}
\end{figure*}

\begin{figure*}[p]
    \centering
    \includegraphics[width=\textwidth]{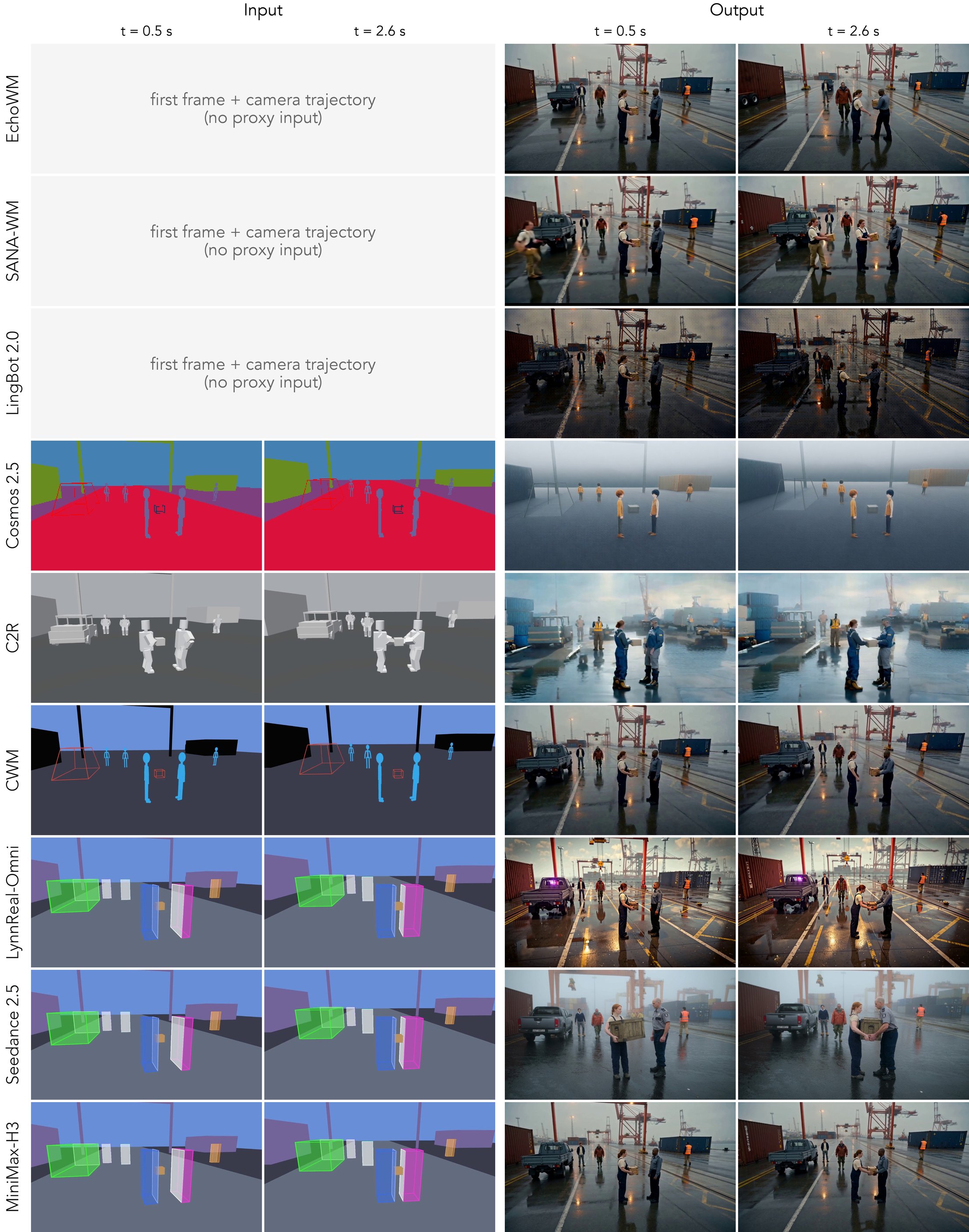}
    \caption{\textbf{\tocheck{All methods on one scene.}} \tocheck{Unverified-FF, handoff scene: a courier hands a parcel to a receiver between 1.0 and 2.6\,s. For each method, the conditioning input (left pair) and the generated output (right pair) at 0.5 and 2.6\,s. Camera-conditioned world models receive frame~0 of the MiniMax-H3 output and the camera trajectory but no proxy; Cosmos-Transfer2.5 receives semantic segmentation (shown) and depth from the CWM proxy representation; CWM additionally receives depth and frame~0 of the MiniMax-H3 output.}}
    \label{fig:handoff_all}
\end{figure*}

\begin{figure*}[p]
    \centering
    \includegraphics[width=\textwidth]{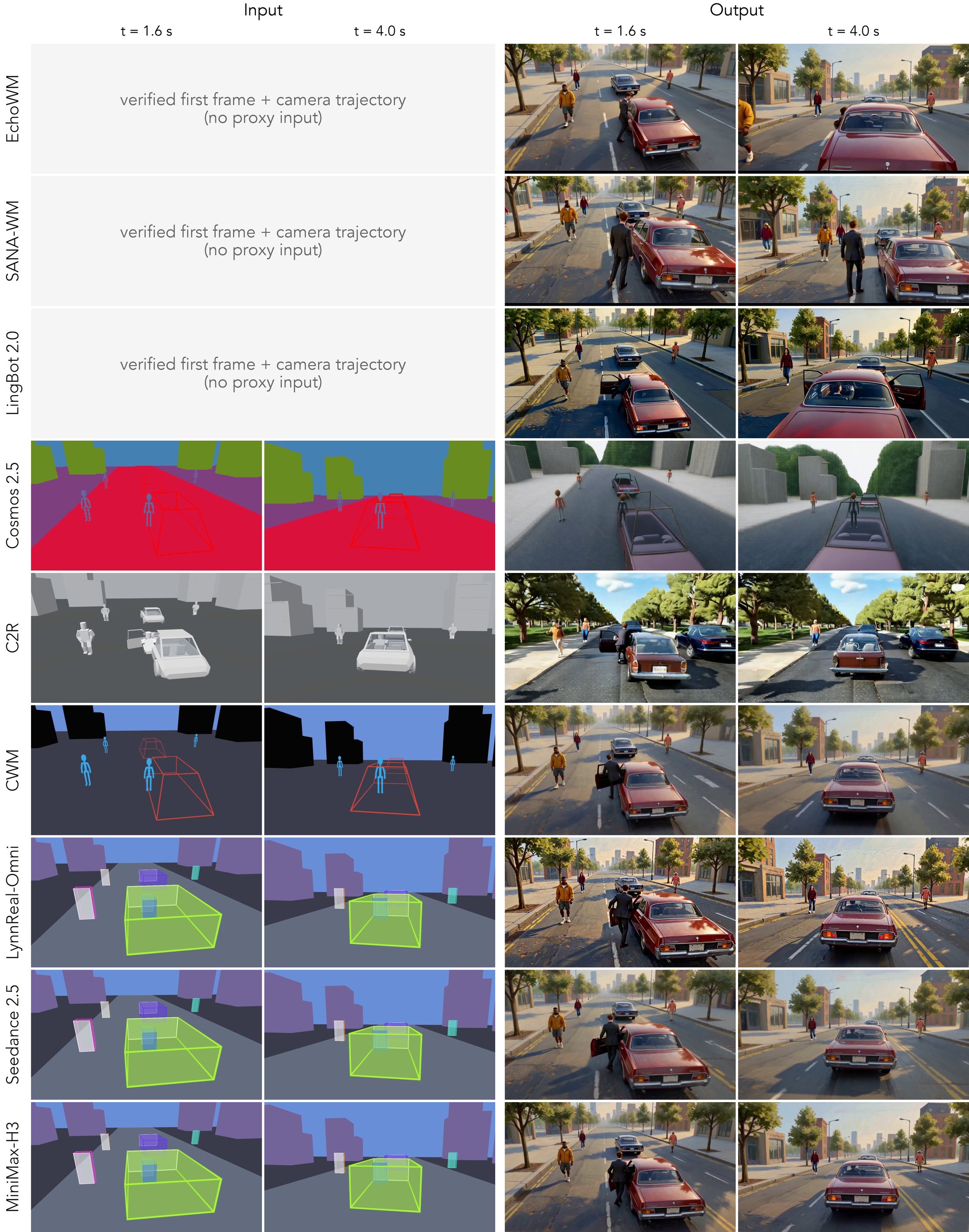}
    \caption{\textbf{\tocheck{All methods on one scene: entering a car.}} \tocheck{Verified-FF, city scene, follow camera: the protagonist walks to a parked car, gets in (1.1--2.2\,s) and drives away (from 2.6\,s); frames at 1.6 and 4.0\,s. Methods that accept a first frame also receive the verified first frame. EchoWM and SANA-WM leave the protagonist outside the car, and LingBot~2.0 opens the door but the car does not leave, whereas the proxy-conditioned methods drive the car away as recorded.}}
    \label{fig:allm_car}
\end{figure*}

\begin{figure*}[p]
    \centering
    \includegraphics[width=\textwidth]{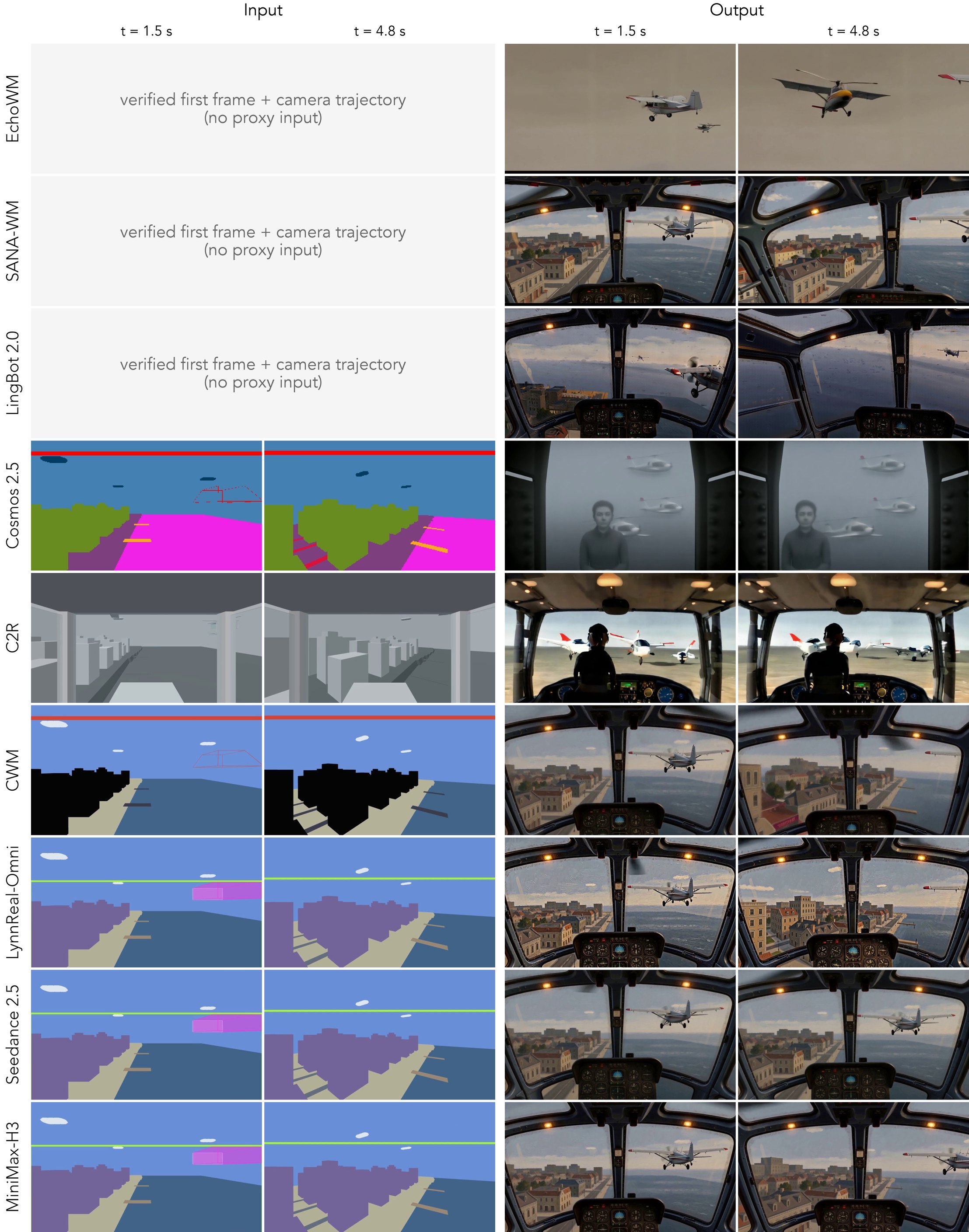}
    \caption{\textbf{\tocheck{All methods on one scene: flying a helicopter.}} \tocheck{Verified-FF, first-person view from a helicopter cockpit that ascends (0--2.3\,s), flies forward (2.3--4.1\,s) and descends over a harbour town, with an aircraft ahead; frames at 1.5 and 4.8\,s. Methods that accept a first frame also receive the verified first frame. EchoWM drops the cockpit and shows the aircraft from outside; Cosmos-Transfer2.5 renders a person facing the camera and C2R a pilot seen from behind, so neither keeps the prescribed first-person view.}}
    \label{fig:allm_heli}
\end{figure*}

\begin{figure*}[p]
    \centering
    \includegraphics[width=\textwidth]{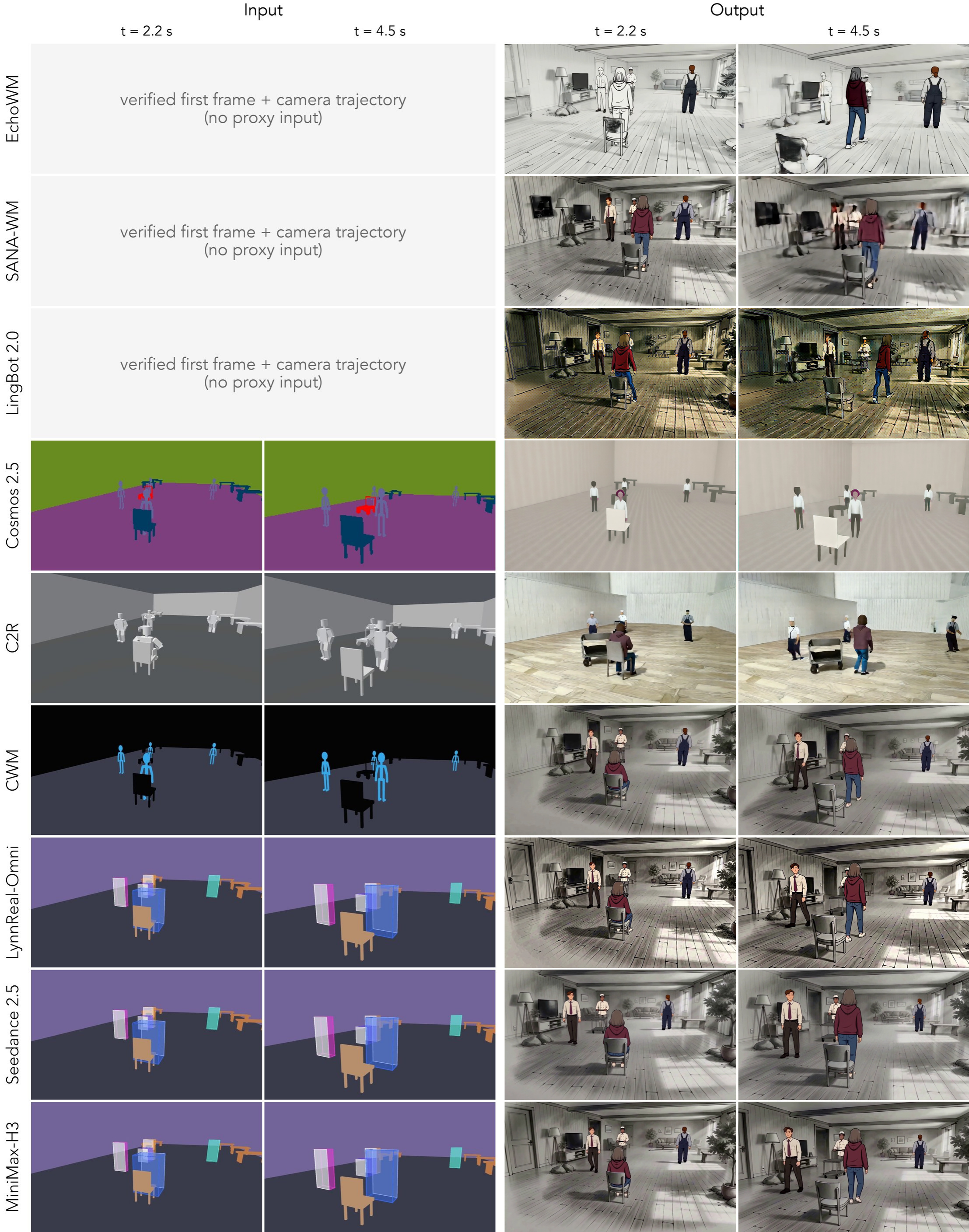}
    \caption{\textbf{\tocheck{All methods on one scene: sitting down and standing up.}} \tocheck{Verified-FF, indoor scene, follow camera: the protagonist sits down on a chair (0.6--1.5\,s), stays seated, and stands up again (3.0--3.9\,s); frames at 2.2\,s (seated) and 4.5\,s (standing). Methods that accept a first frame also receive the verified first frame. EchoWM, SANA-WM and LingBot~2.0 show the protagonist standing at 2.2\,s, when the recorded timeline prescribes a seated pose.}}
    \label{fig:allm_sit}
\end{figure*}

\begin{figure*}[h]
    \centering
    \includegraphics[width=\textwidth]{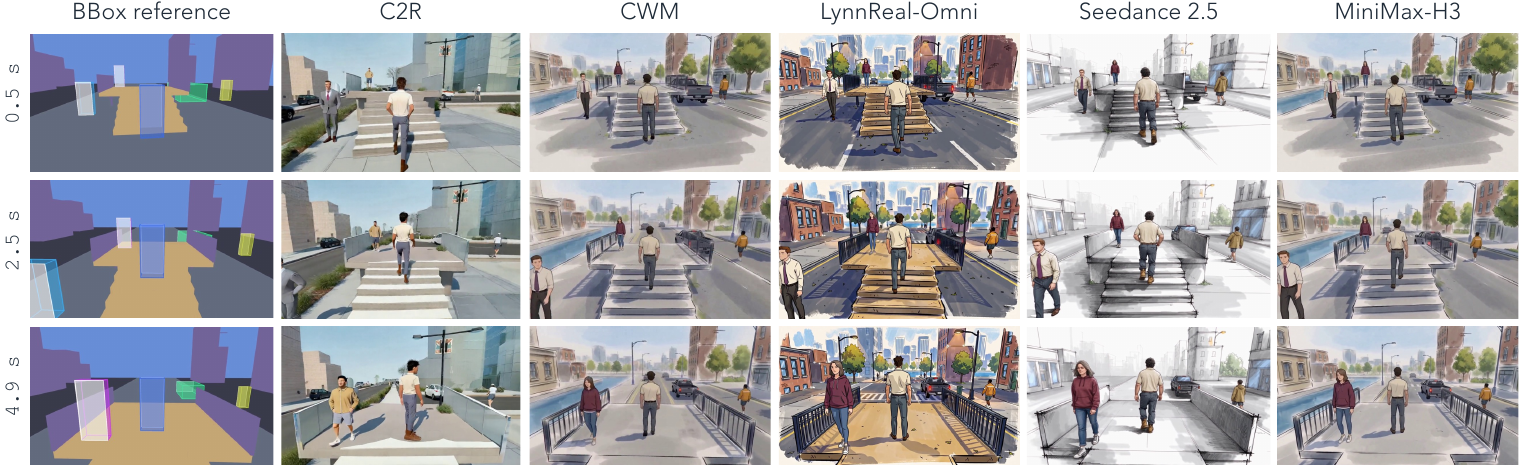}
\caption{\textbf{Style and trajectory without a first frame.} Stair-climbing scene, follow camera, sampled at 0.5, 2.5 and 4.9\,s; the text prescribes a graphic, non-photorealistic marker-sketch style. LynnReal-Omni, Seedance~2.5 and MiniMax-H3 render this style with varying fidelity, and CWM inherits it from MiniMax-H3's frame~0; C2R renders the scene photorealistically although its prompt requests the marker-sketch style. All methods follow the proxy as the protagonist climbs toward the upper landing (pair IoU $0.52$--$0.61$).}
    \label{fig:qual_nofirstframe}
\end{figure*}

\begin{figure*}[h]
    \centering
    \includegraphics[width=\textwidth]{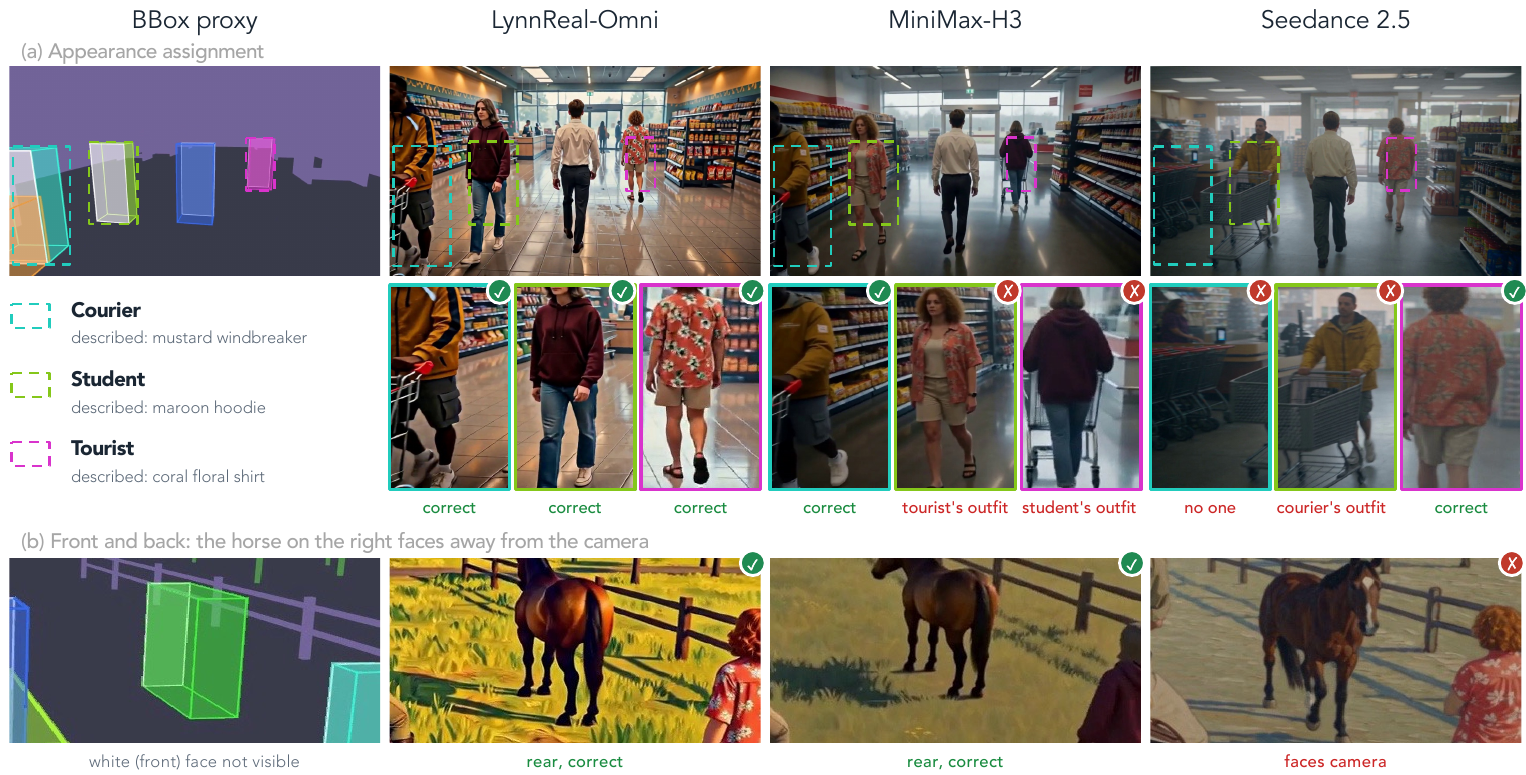}
    % \vspace{-20pt}
\caption{\textbf{Appearance and facing errors without a first frame.} (a) Supermarket scene at 3.0\,s: dashed boxes are the engine reference boxes of a courier, a student and a tourist, each with a described outfit; the zoomed crops show what each method renders inside each box. MiniMax-H3 exchanges the student's and the tourist's outfits; Seedance~2.5 puts the courier's outfit at the student's position. (b) Horseback scene at 0.5\,s: the horse on the right faces away from the camera; Seedance~2.5 renders it facing the camera.
% \zhixiang{It's more like a limitation discussion}
}
    \label{qual_nofirstframe_identity}
\end{figure*}

\tocheck{Figure~\ref{fig:dv200_firstframe_qual} illustrates entity control (T1) of Section~\ref{sec:analysis}, and Figure~\ref{fig:metric_illustration} illustrates Interaction Success Rate and Logic-Render Alignment on one scene.} \tocheck{Figures~\ref{fig:handoff_all}--\ref{fig:allm_sit} show the inputs and outputs of nine methods on four scenes (AlayaWorld~v1.1 omitted).} \tocheck{Figures~\ref{fig:qual_nofirstframe} and~\ref{qual_nofirstframe_identity} show qualitative examples from the Unverified-FF setting discussed in Section~\ref{sec:analysis}.}

% \noindent \textbf{Without a verified first frame, models still render the described world in alignment with the proxy.}
% Current models do not need an appearance reference to render a coherent scene. From the proxy and the text alone, they synthesize the scene in the prescribed style and keep it aligned with the proxy: in Figure~\ref{fig:qual_nofirstframe}, LynnReal-Omni, Seedance~2.5 and MiniMax-H3 reproduce the requested marker-sketch style with varying fidelity, CWM inherits it from MiniMax-H3's frame~0 and C2R renders the scene photorealistically, while in every method the protagonist climbs toward the upper landing at the prescribed times (pair IoU $0.52$--$0.61$). What the first frame would otherwise fix, however, is left to the model: which way each entity faces and which described appearance belongs to which person.

\noindent \textbf{Without a verified first frame, models still render the described world in alignment with the proxy.}
Current models do not need an appearance reference to render a coherent scene. From the proxy and the text alone, most of them synthesize the scene in the prescribed style and keep it aligned with the proxy: in Figure~\ref{fig:qual_nofirstframe}, LynnReal-Omni, Seedance~2.5 and MiniMax-H3 reproduce the requested marker-sketch style with varying fidelity, CWM inherits it from MiniMax-H3's frame~0, and C2R renders the scene photorealistically although its prompt requests the same style, while in every method the protagonist climbs toward the upper landing at the prescribed times (pair IoU $0.52$--$0.61$). What the first frame would otherwise fix, however, is left to the model: which way each entity faces and which described appearance belongs to which person.

% \noindent \textbf{Without a verified/prescribed first frame, models still render the described world in alignment with the proxy.}
% Current models do not need an appearance reference to render a coherent scene. From the proxy and the text alone, they synthesize the scene in the prescribed style and keep it aligned with the proxy: in Figure~\ref{fig:qual_nofirstframe}, every method renders reproduce the requested marker-sketch style\TODO{Fix the grammar (``renders reproduce''); this also contradicts the caption of Figure~\ref{fig:qual_nofirstframe}, where C2R renders the scene photorealistically and CWM inherits the style from MiniMax-H3's frame~0.} with varying fidelity, and the protagonist climbs toward the upper landing at the prescribed times (pair IoU $0.52$--$0.61$). What the first frame would otherwise fix, however, is left to the model: which way each entity faces and which described appearance belongs to which person.

% \section{Additional Results}

\end{document}